\documentclass{article}

\usepackage{microtype}
\usepackage{graphicx}
\usepackage{subfigure}
\usepackage{booktabs} 
\usepackage{xspace}
\usepackage{xcolor}
\usepackage{tcolorbox}
\tcbuselibrary{skins}
\usepackage{amsmath}
\usepackage{amssymb}
\usepackage{wrapfig}

\usepackage{hyperref}

\definecolor{PreRoPE}{HTML}{B5547A}
\definecolor{PostRoPE}{HTML}{287FA8}

\definecolor{CommentGreen}{HTML}{427A46}
\newcommand{\algcomment}[1]{\hfill{\color{CommentGreen}\footnotesize // #1}}

\usepackage{hyperref}

\usepackage[accepted]{mlsys2025}
\makeatletter
\renewcommand{\Notice@String}{}
\makeatother

\mlsystitlerunning{ReelCache: An Efficient KV Cache Compression and Management System for Multi-Round Video Generation}

\definecolor{DaiyangPurple}{HTML}{9B30FF}

\newcommand{\systemname}{\textsf{ReelCache}\xspace}

\definecolor{mygray}{gray}{0.9}

\newtcolorbox{observationbox}[1]{
  enhanced,
  colback=gray!7,
  colframe=black!80,
  boxrule=0pt,
  leftrule=3pt,
  sharp corners,
  boxsep=4pt,
  left=4pt,
  right=4pt,
  top=4pt,
  bottom=4pt,
  title={\textbf{#1}},
  coltitle=black,
  attach title to upper={\quad}
}

\newcommand{\squishlist}{
    \begin{list}{$\bullet$}
        { \setlength{\itemsep}{0pt}      \setlength{\parsep}{0pt}
            \setlength{\topsep}{0.5pt}       \setlength{\partopsep}{0pt}
            \setlength{\listparindent}{-2pt}
            \setlength{\itemindent}{-5pt}
            \setlength{\leftmargin}{0.7em} \setlength{\labelwidth}{0em}
            \setlength{\labelsep}{0.2em} } }
    
\newcommand{\squishend}{
\end{list}  }

\definecolor{asparagus}{rgb}{0.55, 0.71, 0.0}

\newcounter{squishnum} 

\newcommand{\squishenum}{
    \begin{list}{\arabic{squishnum}.} 
        { \usecounter{squishnum}      
          \setlength{\itemsep}{0pt}
          \setlength{\parsep}{0pt}
          \setlength{\topsep}{0.5pt}
          \setlength{\partopsep}{0pt}
          \setlength{\listparindent}{-2pt}
          \setlength{\itemindent}{-5pt}
          \setlength{\leftmargin}{0.7em} 
          \setlength{\labelwidth}{0em}
          \setlength{\labelsep}{0.2em} } }

\newcommand{\squishenumend}{
\end{list}  }

\begin{document}

\twocolumn[
\mlsystitle{ReelCache: An Efficient KV Cache Compression and Management System for Multi-Round Video Generation}

\mlsyssetsymbol{equal}{*}

\begin{mlsysauthorlist}
 \mlsysauthor{Tianyu Wang}{Pitt}
 \mlsysauthor{Daiyang Yu}{Pitt}
 \mlsysauthor{Zhiwei Ren}{Pitt}
 \mlsysauthor{Junbo Li}{Pitt} 
 \mlsysauthor{Jitian Zhang}{Google} 
 \mlsysauthor{Longfei Shangguan}{Pitt} 
\end{mlsysauthorlist}
 
\mlsysaffiliation{Pitt}{University of Pittsburgh, PA, USA}
\mlsysaffiliation{Google}{Google, CA, USA}
\mlsyscorrespondingauthor{Tianyu Wang}{tianyu.wang@pitt.edu}
 



\mlsyskeywords{Machine Learning, MLSys}


\begin{abstract}
Autoregressive video diffusion models produce large KV caches during multi-round video generation. Keeping all these KV chunks would consume substantial memory, while rebuilding discarded caches would significantly delay video chunk generation. This paper presents \systemname, a KV cache compression system for efficient multi-round, long video generation. \systemname combines an intra-chunk compressor with an inter-chunk cache manager. The compressor applies low-rank decomposition across channels and video coding to the resulting coefficients, predicting the decomposition rank per transformer layer online to meet retained-energy targets. The cache manager shares basis directions across consecutive chunks and organizes chunk states in a tree, allowing generation branches to share historical state while supporting restoration for resume, rewind, and fork.
Experiments across five video generation models show that \systemname reduces KV storage by 14.46--21.13$\times$ relative to baselines and accelerates next-chunk generation by 1.25--3.36$\times$, with less than 1\% drop in six representative VBench metrics.
\textit{GitHub: \url{https://github.com/Tianyu9748/ReelCache}}
\end{abstract}
]



\printAffiliationsAndNotice{} 

\section{Introduction}
\label{s:introduction}

Video generation tools such as Google Veo~\cite{veo3_techreport}, ByteDance Seedance~\cite{seedance2026seedance}, and Adobe Firefly~\cite{adobe_firefly_video2024} empower users to 
create video clips from text descriptions or reference
images~\cite{wan,sora}. These tools make it possible to produce professional visual content without extensive animation experience, significantly lowering the entry barrier to video creation. When creating a longer video, a user may start with a short clip and extend it through successive rounds of generation. Each continuation needs to build on the preceding content, preserving the appearance of recurring characters and
objects and the continuity of the scene.

Autoregressive video diffusion models support this continuity by generating video chunk by chunk, with each new chunk conditioned on preceding content~\cite{causvid,selfforcing,longlive,magi1}.
To avoid repeatedly processing preceding chunks, these systems cache their attention keys and values (KV). When generating a new chunk, the model attends directly to the cached KV rather than recomputing these representations from the historical video latents. However, retaining these per-layer KV representations
consumes substantial GPU memory. For example, in our benchmark with LongLive-1.0-1.3B at 1080p resolution, the KV cache for five seconds of video occupies 45~GB memory (\S\ref{ss:historical_kv_cost}). Keeping multiple such histories would quickly exceed GPU memory capacity.

Multi-round video creation makes this problem more demanding (\S\ref{ss:interactive_reuse}). When generating a longer video, users typically review a few seconds or tens of seconds of generated video before deciding how to proceed. For example, they may adjust the prompt for subsequent chunks, return to an earlier point to replace an unsatisfactory segment, or explore alternative continuations. These operations can revisit generation points whose KV is no longer in the active cache. The system should therefore either retain extended KV for later reuse which is memory intensive, or reconstruct it when needed which comes with significant computing delay.


In this paper, we present \systemname, a KV-cache compression and management system for multi-round, long video generation. \systemname compresses each chunk's KV as it is generated, reducing the storage needed to retain long generation histories. When a user edits an earlier chunk, it retrieves and restores the compressed KV of the required preceding chunks, avoiding costly recomputation from historical video latents. It uses CPU/NVENC for video encoding and dedicated GPU video decoders for coefficient decoding, limiting interference with ongoing video generations. 

\systemname's design is guided by three key findings from our characterization of video KV caches: complementary channel
and spatiotemporal redundancy, rank requirements that vary
across prompts, layers, and models, and shared low-rank
structure across consecutive chunks
(\S\ref{s:idea_challenge}).


Our first finding is that video KV contains complementary redundancy across channels and spatial and temporal positions (\S\ref{ss:motivation_svdcodec}). 
Prior work has used low-rank decomposition and video codecs
for LLM KV compression~\cite{shadowkv,palu,llm265,kvcodec}. Our characterization of video KV reveals how these techniques can work well together: low-rank decomposition reduces channel redundancy, while the resulting coefficient channels retain spatial and temporal correlations that a video codec can further exploit. We also find that, at the same rank, compressing keys before RoPE better preserves subsequent video generation quality than compressing them after RoPE. These findings guide us to design an intra-chunk compressor that factorizes pre-RoPE keys and values, then encodes their coefficient channels with a video codec (\S\ref{ss:design_intra_chunk}). The combination reduces storage beyond either method alone.

Nevertheless, our measurements also show that keys and values differ in their sensitivity to compression, and that a fixed retained-energy target requires substantially different
ranks across layers, prompts, and models. A uniform rank (for low-rank decomposition-based compression) therefore either wastes storage on some KV matrices or discards too much information from others (\S\ref{ss:motivation_rank}). Computing a full singular spectrum to select each rank, however, would add substantial computing overhead. 
In \systemname we propose a small-rank one-time energy measurement coupled with a learned rank-energy relationship to predict the required rank, then verify the result against the
energy target (\S\ref{ss:design_adaptive_rank}). This makes per-matrix rank selection possible with significantly lower computation overhead.

Our second finding is that although neighboring video chunks' raw KV tensors differ substantially, their channel subspaces often remain similar. This creates an opportunity to share a common basis among consecutive KV caches while storing separate coefficients for each chunk (\S\ref{ss:motivation_sharingbasis}). 
To exploit this opportunity, \systemname checks whether the previous chunk's basis captures enough energy in the new chunk's KV matrix. If so, the new chunk stores only its own coefficients and references the shared basis. Otherwise, \systemname adds directions to capture the residual, or fits a new basis if doing so requires less storage. To support multi-round video editing, we further propose an inter-chunk manager that organizes chunks into a \emph{chunk state tree}, where branches represent alternative continuations and share their common history. Each node stores the chunk's encoded coefficients and references to its shared basis. Coefficients are encoded separately for each chunk, allowing it to be restored using its referenced basis without decoding earlier chunks' coefficients (\S\ref{ss:design_inter_chunk}).

We implement \systemname in  21K LoC and integrate it into SGLang-Diffusion~\cite{sglang}. The head-to-head comparison against two baselines across five video generation models shows that \systemname reduces KV storage by up to 21.13$\times$ and accelerates next chunk generation by up to 3.36$\times$, without noticeable video quality degradation.

This paper makes the following contributions:

\squishlist{}
    \item We characterize the compression opportunities in video KV caches across channels, spatial and temporal positions, and consecutive chunks through benchmark studies. 
    \item Guided by the findings from these benchmarks, we design \systemname, an efficient KV-cache compression system that exploits both inner- and inter-chunk KV redundancy.
    \item We integrate \systemname into SGLang-Diffusion and demonstrate it consistently outperforms baselines. 
\squishend{}

\begin{figure*}[t]
\centering
\begin{minipage}[t]{0.32\textwidth}
  \vspace{0pt}
  \centering
  \includegraphics[width=\linewidth]{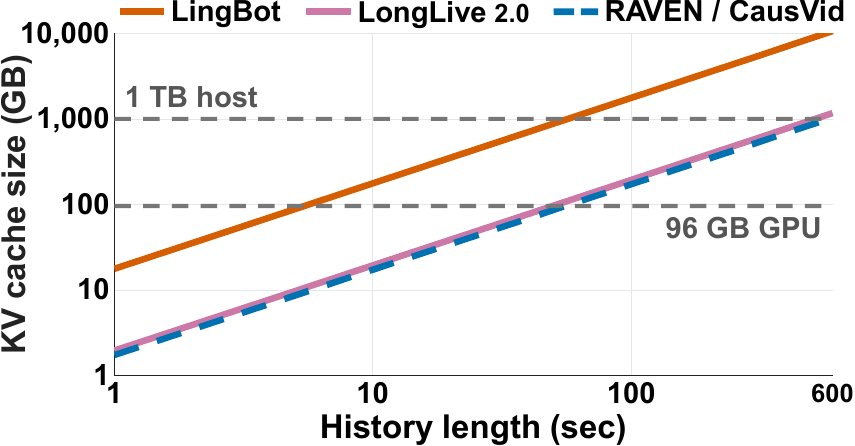}
  \vspace{-8mm}
  \caption{Size of the KV and the longest history that fits in one RTX PRO 6000 with 96~GB memory or in 1~TB of host memory.}
  \label{fig:kv_storage}
\end{minipage}\hfill
\begin{minipage}[t]{0.32\textwidth}
  \vspace{0pt}
  \centering
  \includegraphics[width=\linewidth]{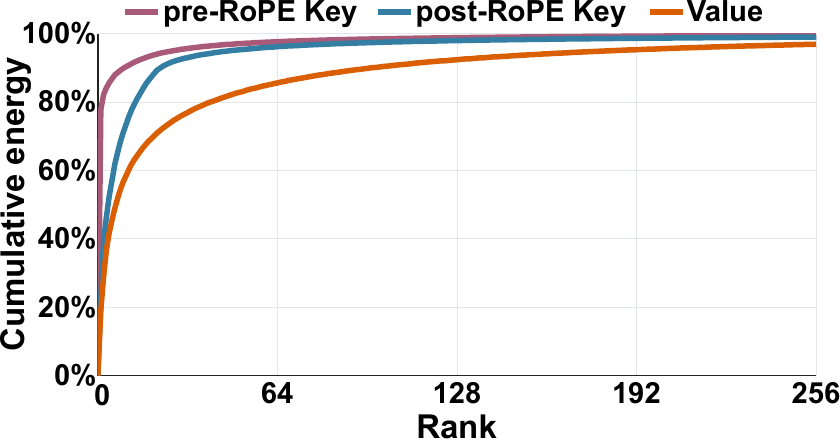}
   \vspace{-8mm}
  \caption{Video KV is low-rank across the channel dimension:
  a small basis captures almost all of the matrix's energy.}
  \label{fig:lowrank}
\end{minipage}\hfill
\begin{minipage}[t]{0.32\textwidth}
  \vspace{0pt}
  \centering
  \includegraphics[width=\linewidth]{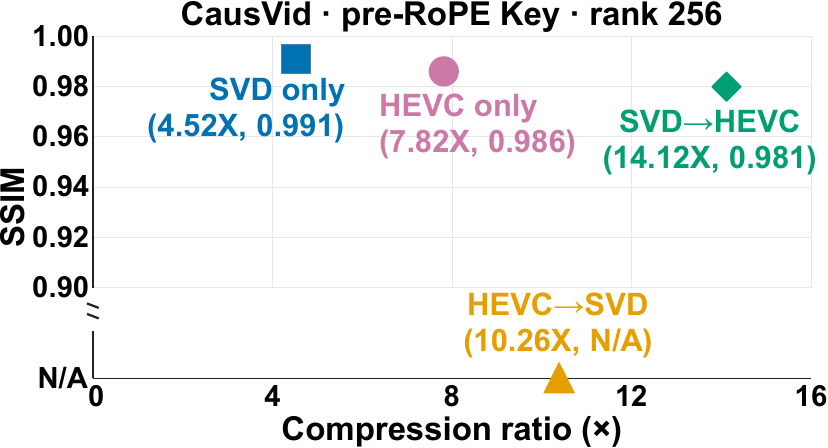}
   \vspace{-8mm}
  \caption{Compression ratio and SSIM of pre-RoPE Key
  compression (CausVid generates one chunk with SVD rank at 256).}
  \label{fig:compression_pipeline}
\end{minipage}
\vspace{-6mm}
\end{figure*}

\vspace{-4mm}

\section{Related Works}
\label{s:related_works}


\textbf{KV-cache compression for LLMs.}
KV cache takes significant memory on long-context LLM inference. Prior works can be divided into the following categories: 
token eviction/discarding keeps only a budget of ``important'' tokens and drops the rest~\cite{h2o,streamingllm,snapkv,pyramidkv};
quantization lowers bits per element~\cite{kivi,kvquant,gear};
low-rank/structural methods project tensors, which lie in a low-rank subspace, onto a small basis~\cite{palu, shadowkv}. 
Close to \systemname, LLM.265~\cite{llm265} shows H.264/H.265  codec can be leveraged to quantize LLM KV cache to $\sim\!3$ bits, thereby reducing storage overhead. KVCodec~\cite{kvcodec} further adds a codec-friendly layout and a GPU-native NVENC/NVDEC pipeline for lossless remote KV reuse and migration.

\systemname differs from these works in two aspects.
First, these works optimize LLM decode bandwidth per step subject to accuracy, whereas \systemname optimizes persistent bytes per branchable snapshot subject to reproducibility.
Second, the codec line runs on the raw tensor and never removes the cross-channel redundancy that, for video KV that \systemname targets at, is the larger one (\S\ref{s:motivation}). 


\noindent\textbf{Inter-step activation caching for faster diffusion inference.} Prior works~\cite{activation_cache_lv2025fastercache,activation_cache_pyramid_kv} exploit similarity between intermediate activations across denoising steps to avoid repeating selected computations during generation.
\systemname addresses a different but complementary source of overhead: recovering the historical KV required to generate the next chunk. Thus, inter-step activation caching reduces computation within the current chunk's denoising process, whereas \systemname reduces the cost of making historical KV available.
Detailed comparison is provided in Appendix~\ref{app:cache_activations}.
We also review efficient and parallel video generation systems in Appendix~\ref{s:related_efficient_parallel_video_infer}.

\vspace{-3mm}
\section{Background and Motivation}
\label{s:motivation}



In this section, we first characterize video generation (\S\ref{ss:interactive_reuse}) and then explain how retaining KV caches for the multi-turn video generation leads to prohibitive memory footprint (\S\ref{ss:historical_kv_cost}). 
The illustration of video denoising and KV conditioning in autoregressive video diffusion is listed in Appendix~\ref{ss:avgp}.



\subsection{Characterizing Long Video Generation}
\label{ss:interactive_reuse}

While retaining only the latest or active KV window, as  LongLive~\cite{longlive}, supports efficient forward video generation, it is insufficient for long video generation tasks. In these tasks, video creation often involves multiple rounds of generation and revision.
For example, after generating a one-minute video of a person entering a cafe, ordering a drink, and sitting by the window, a user may decide that the person should instead walk toward the table immediately after entering, around the 5th second.
The video generation system should then return to that early point, preserve the first 5s footage, and generate a new continuation.
At this point, however, the active KV window would contain only the context near the end of the original one-minute video, while the KVs needed to continue from the 5th second may have already been evicted long ago.
As a result, without a saved copy of the required KV state, the server has to reconstruct it before generating the revised sequence, imposing prolonged waiting times on users.

\begin{figure}[t]
\centering
\includegraphics[width=\columnwidth]{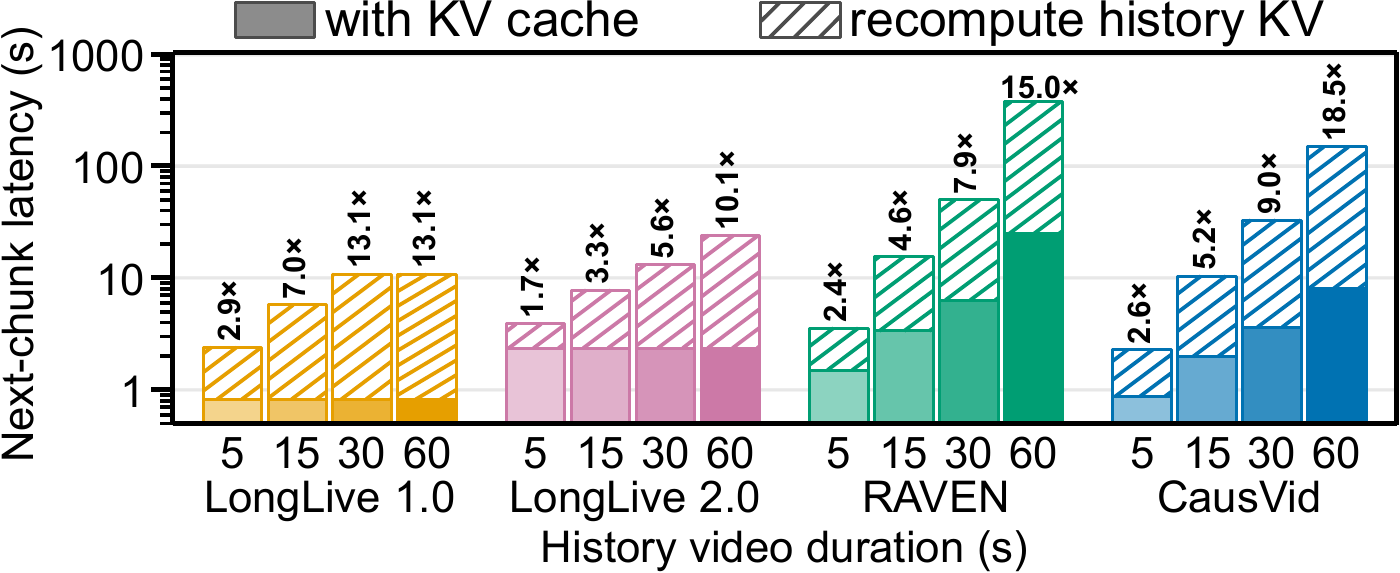}
\vspace{-9mm}
\caption{Next-chunk latency with history KV cache resident and recomputed from latents across models and history durations on an RTX PRO 6000. Text encoding and VAE decoding are excluded.}
\vspace{-6mm}
\label{fig:recompute}
\end{figure}

Figure~\ref{fig:recompute} quantifies this overhead on an RTX PRO 6000 GPU.
We observed that the overhead of rebuilding historical KV grows sharply with video history length across all models.
Compared with reusing resident KV (on GPU), rebuilding the cache from latents raises total next-chunk latency to $1.7$--$2.9\times$ the baseline for a 5s history.
For a 60s history, this latency gap increases to $10.1$--$18.5\times$, with slowdowns of $13.1\times$, $10.1\times$, $15.0\times$, and $18.5\times$ for LongLive 1.0 \cite{longlive}, LongLive 2.0 \cite{longlive2.0}, RAVEN \cite{raven}, and CausVid \cite{causvid}, respectively.


\subsection{The Excessive Cost of Retaining Historical KV}
\label{ss:historical_kv_cost}

A straightforward way to avoid repeated reconstruction and delay is to retain all previously generated KV chunks. However, preserving this KV history incurs a substantial memory cost. Video represents each frame as a spatial grid of tokens (e.g., 16$\times$16 pixels), and each chunk spans several such frames, producing a large number of tokens even for a short sequence. Each token further requires keys and values to be stored across transformer layers. Retaining all KV therefore makes storage a major obstacle to historical state reuse.



Figure~\ref{fig:kv_storage} shows the estimated FP16 historical KV cache for four models, assuming 24~FPS video.
Each second of generated video adds approximately 1.95~GB of KV for LongLive 2.0 and 1.73~GB for CausVid.
For LingBot (a 13.6B model at 720p), this number grows to 17.0~GB per second.
Consequently, even if all 96~GB of an RTX PRO 6000 were dedicated to historical KV, it would hold less than one minute of history for the smaller models and only 5.6s for LingBot. Note that these are optimistic capacity bounds. In practice, model weights and inference activations would compete with KV cache for the same GPU memory. 

\noindent\textbf{How about offloading to the host memory?}
Historical KV can instead be offloaded to host memory, as in LLM serving systems~\cite{lee2024infinigen,liu2025lmcache}.
However, its large footprint remains a capacity concern. For instance, as shown in Figure~\ref{fig:kv_storage}, a dedicated 1~TB host-memory budget holds approximately 8.6 minutes of LongLive 2.0 history or only 59 seconds of LingBot history, before accounting for other uses of host memory.

Moreover, due to the limited PCIe bandwidth, migrating KV cache from the host memory or hard disk to GPU still comes with significant delay. For example, for CausVid with 30s of history, the required KV occupies 51.8~GB. Resident KV on the GPU yields the next chunk generation in 3.64s. In contrast, migrating KV from host memory or SSD can take 2.46s and 11.23s, respectively.


\vspace{-2mm}
\section{Basic Idea and Design Rationale}
\label{s:idea_challenge}





In this paper, we revisit a basic storage strategy for multi-turn video generation: {\it compress historical KV when retaining it for future reuse, and decompress the required state on demand}. This reduces both the storage footprint of historical KV cache and the amount of data transferred between main memory and GPU when restoring an earlier generation point. However, realizing this basic idea requires answering three core questions:

(i) {\it How to exploit the redundancy within each chunk's KV?}
\\
(ii) {\it How much to compress each KV matrix while preserving generation quality?}
\\
(iii) {\it Whether consecutive chunks share structure that can further reduce storage and compression overhead?}

We investigate these questions through the measurements below, which establish the rationale for our design.

\noindent\textbf{Measurement setup.}
We characterize both the structure of historical KV and the effect of its compression on subsequent generation.
Because reconstructed KV conditions future chunks, we assess compression fidelity using the Structural Similarity Index Measure (SSIM) of the decoded continuation relative to an uncompressed-cache reference generated with the same prompt and random seed.
We use VBench~\cite{vbench} to assess generated-video quality.
Unless otherwise noted, experiments use CausVid~\cite{causvid} at a fixed resolution of $832{\times}480$ pixels.



\begin{figure}[t]
\centering
\includegraphics[width=\columnwidth]{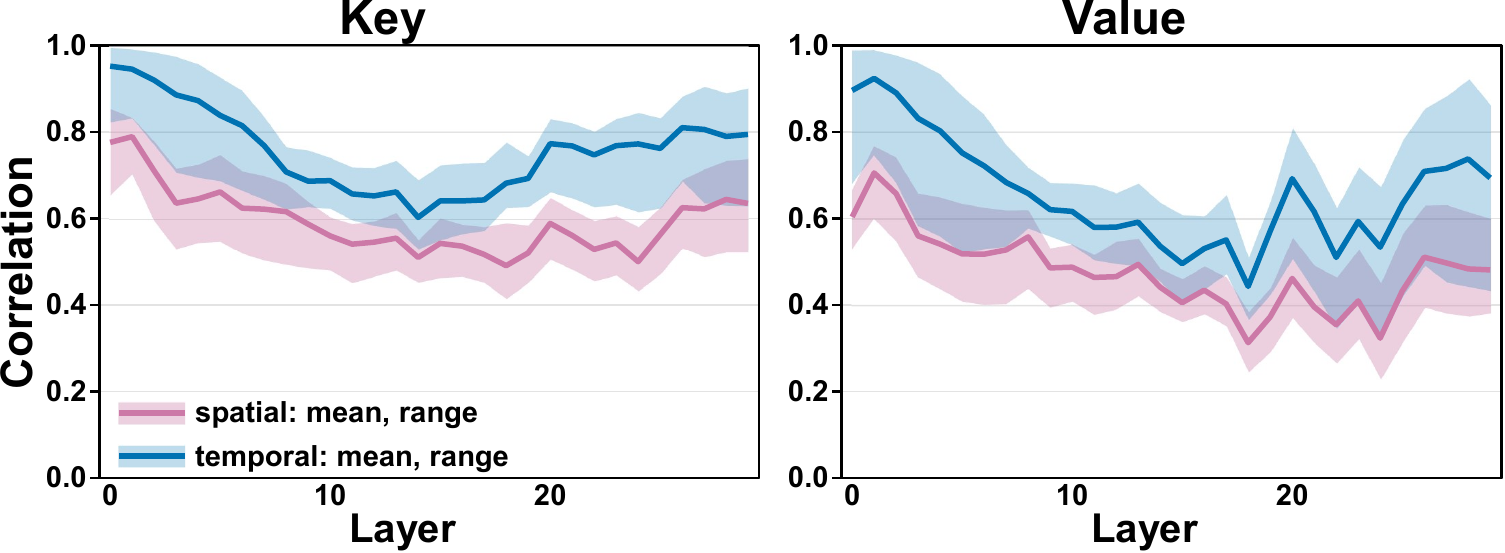}
\vspace{-9mm}
\caption{Per-layer spatial and temporal correlation (mean, max, and min) within each pre-RoPE Key and Value channel of CausVid.}
\label{fig:kv_spatialtemporal_redundancy}
\vspace{-6mm}
\end{figure}

\vspace{-2mm}
\subsection{Complementary Redundancy Within a Chunk}
\label{ss:motivation_svdcodec}

\begin{observationbox}{Observation One:}
Video KV cache exhibits complementary channel and spatiotemporal redundancy.
The measurement motivates us to combine low-rank factorization with video codec encoding, using the pre-RoPE representation for the Key.
\end{observationbox}


\noindent$\bullet$ {\it Across channels}, we found that the KV matrices exhibit a concentrated singular spectrum, indicating that many channel features can be represented by a smaller set of shared directions.
Figure~\ref{fig:lowrank} shows a concrete example: despite its 1536-dimensional channel representation, 96 singular directions (6.25\%) capture 98\% of the pre-RoPE Key's energy.

\noindent$\bullet$ {\it Within each channel}, KV matrices also exhibit extensive spatial and temporal correlations, as shown in Figure~\ref{fig:kv_spatialtemporal_redundancy}.
It is understandable that nearby regions within each frame often depict similar content, while consecutive frames share much of the same scene, with changes concentrated around moving objects.
These spatial-temporal redundancies are reflected in the corresponding KV features, which create redundancy across spatial positions and neighboring frames. 

These two degrees of redundancy motivate us a two-stage compression pipeline: a): apply SVD to reduce redundancy across channels, representing the KV tensor with fewer coefficient channels and a compact basis, and b): use a video codec to compress spatial and temporal redundancy within these coefficient channels.
Figure~\ref{fig:compression_pipeline} illustrates the potential of combining these mechanisms.
For the pre-RoPE Key, HEVC (a.k.a., H.265, a representative video codec) alone achieves $7.82\times$ compression with a continuation SSIM of 0.986, while rank-256 SVD achieves $4.52\times$ with SSIM 0.991.
Applying HEVC to the SVD coefficients increases compression to $14.12\times$, with SSIM slightly drops to 0.981.

\begin{table}[t]
\centering
\small
\caption{Next-chunk SSIM with history Key compressed by SVD
before or after RoPE (one history chunk, Value uncompressed).}
\vspace{0mm}
\label{tab:prepost}
\begingroup
\setlength{\tabcolsep}{2.9pt}
\renewcommand{\arraystretch}{1.18}
\begin{tabular*}{\columnwidth}{
  @{\extracolsep{\fill}}lcccc@{}
}
\toprule[0.7pt]
& \multicolumn{2}{c}{Rank 64}
& \multicolumn{2}{c}{Rank 256} \\
\cmidrule(lr){2-3}
\cmidrule(lr){4-5}
Model
& \textcolor{PreRoPE}{pre-RoPE}
& \textcolor{PostRoPE}{post-RoPE}
& \textcolor{PreRoPE}{pre-RoPE}
& \textcolor{PostRoPE}{post-RoPE} \\
\midrule[0.4pt]
LongLive 2.0 & \textbf{0.823} & 0.769
             & \textbf{0.924} & 0.885 \\
RAVEN        & \textbf{0.989} & 0.966
             & \textbf{0.998} & 0.996 \\
CausVid      & \textbf{0.970} & 0.918
             & \textbf{0.995} & 0.990 \\
\bottomrule[0.7pt]
\end{tabular*}
\endgroup
\vspace{-6mm}
\end{table}


While the initial results are promising, we also found that the representation being compressed also matters.
RoPE applies position-dependent rotations to the Key, changing its structure across token positions.
Table~\ref{tab:prepost} compares low-rank compression before and after RoPE at matched ranks and compression ratios, leaving the Value uncompressed.
The pre-RoPE representation consistently yields higher continuation SSIM across the tested models, with larger differences at lower ranks.
For example, at rank 64, CausVid achieves SSIM 0.970 with pre-RoPE Key compression versus 0.918 with post-RoPE compression.
These results favor compressing the Key before RoPE and restoring its positional encoding after reconstruction.

\subsection{Heterogeneous Rank Requirements}
\label{ss:motivation_rank}

\begin{observationbox}{Observation Two:}
The SVD rank needed to meet a retained-energy target varies across layers, prompts, and models.
Efficient compression thus requires per-matrix rank selection with low selection overhead.
\end{observationbox}

When applying SVD, we have to determine how many singular directions (i.e., rank) to retain.
A smaller rank compresses more, but also loses more information that may affect subsequent chunk frame generation. Conversely, a larger rank preserves more information but increases storage. We first examine retained spectral energy as a proxy for compression fidelity, and then characterize the rank needed to meet a given energy target.

\begin{figure}[t]
\centering
\includegraphics[width=\columnwidth]{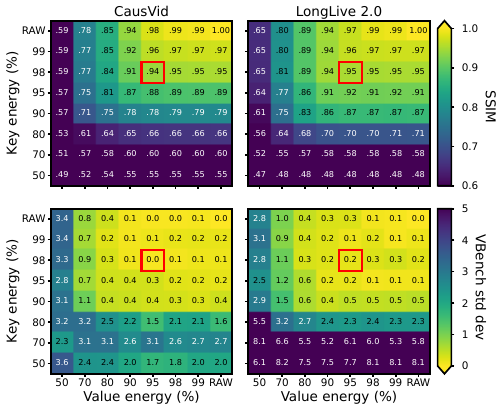}
\vspace{-9mm}
\caption{SSIM and VBench standard deviation relative to the uncompressed cache, for CausVid and LongLive 2.0 across retained Key and Value energy. The red box marks 98\% Key / 95\% Value.}
\vspace{-6mm}
\label{fig:energy_video_heatmap}
\end{figure}

As shown in Figure~\ref{fig:energy_video_heatmap}, we found that the SSIM metric generally improves as more energy is retained, but the Key and Value exhibit different sensitivities.
As the figure shows, on CausVid, retaining 90\%, 95\%, and 98\% of the Value energy (with the Key uncompressed) yields an SSIM of 0.94, 0.98, and 0.99, respectively.
In contrast, retaining the same fractions of the Key energy (with the Value uncompressed) yields an SSIM of only 0.79, 0.89, and 0.95, respectively.
Thus, equal retained-energy fractions do not imply equal effects on the generated video.

Suggested by the above benchmark result, we retain 98\% of the Key energy and 95\% of the Value energy to preserve generated-video quality.
Under joint Key and Value compression, these targets keep VBench scores within 0.2 points of the uncompressed reference on both CausVid and LongLive 2.0, indicating minimal quality degradation on the evaluated benchmark.
We therefore use these targets in the following characterization to examine compression opportunities while maintaining near-reference video quality.


\begin{figure}[t]
\centering
\includegraphics[width=\columnwidth]{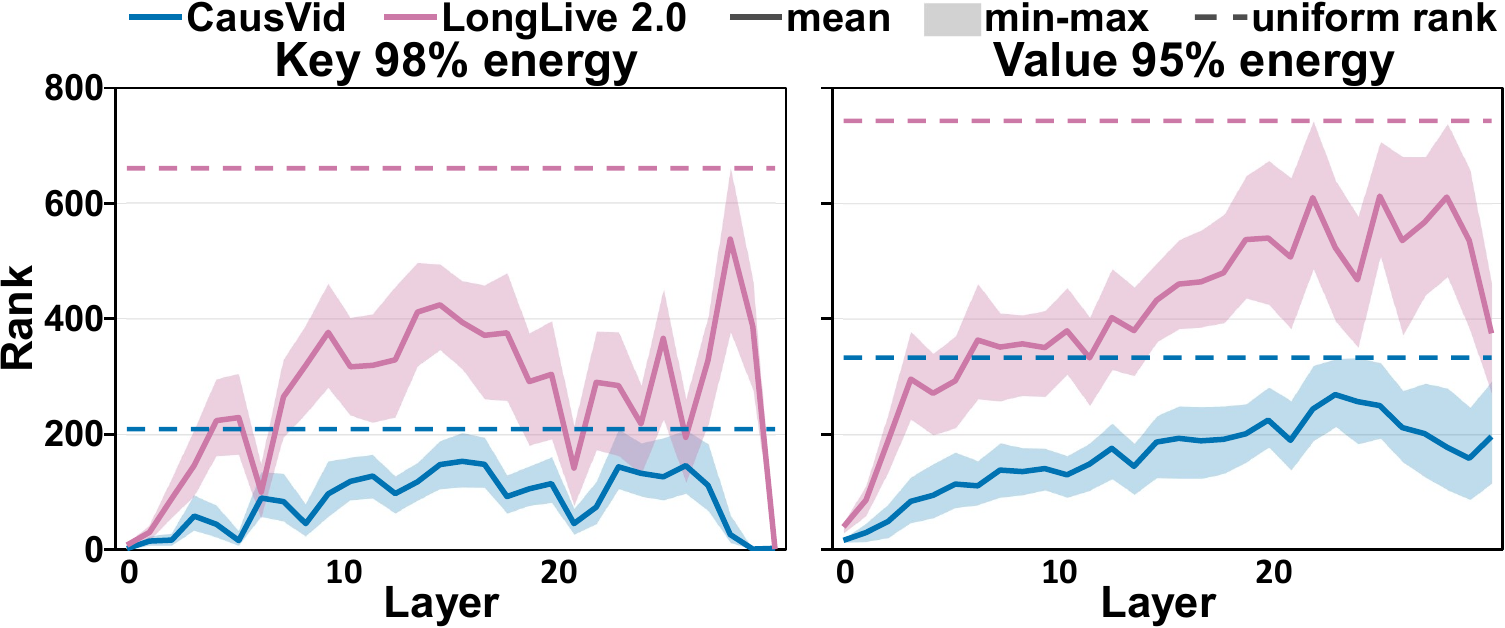}
\vspace{-9mm}
\caption{Smallest rank that retains 98\% of Key energy and 95\% of Value energy for each layer of CausVid and LongLive 2.0. The dashed line is the uniform rank that meets the target everywhere.}
\vspace{-3mm}
\label{fig:rankvar}
\end{figure}

\noindent\textbf{A fixed energy target requires different ranks}.
We then run experiments to explore the rank required to meet the retained-energy target in each layer of
CausVid across the VBench dataset.
Figure~\ref{fig:rankvar} shows substantial variation across layers and prompts.
For CausVid, the mean required rank ranges from 1 to 153 for Key and from 16 to 269 for Value across different layers.
For LongLive 2.0, this mean rank ranges from 1 to 538 for Key and from 39 to 613 for Value.

We also found the required rank differs across models as well. On average, CausVid requires ranks of 83 for Key and 162 for Value out of 1536 channels. LongLive 2.0, on the other hand, requires ranks of 269 and 412 out of 3072 channels.  To meet the targets for every KV, a uniform rank would need to increase to 209 and 333 on CausVid, and 661 and 743 on LongLive 2.0. Such a uniform rank setup would allocate unnecessary singular directions to matrices that admit substantially lower-rank representations, wasting precious storage. Compared with giving each matrix its own rank, it stores 1.21$\times$ more SVD factors on CausVid and 1.06$\times$ more on LongLive 2.0. This variation indicates that we should select the rank separately for each KV matrix in order to maximize the compression ratio.

\begin{figure}[t]
\centering
\includegraphics[width=\columnwidth]{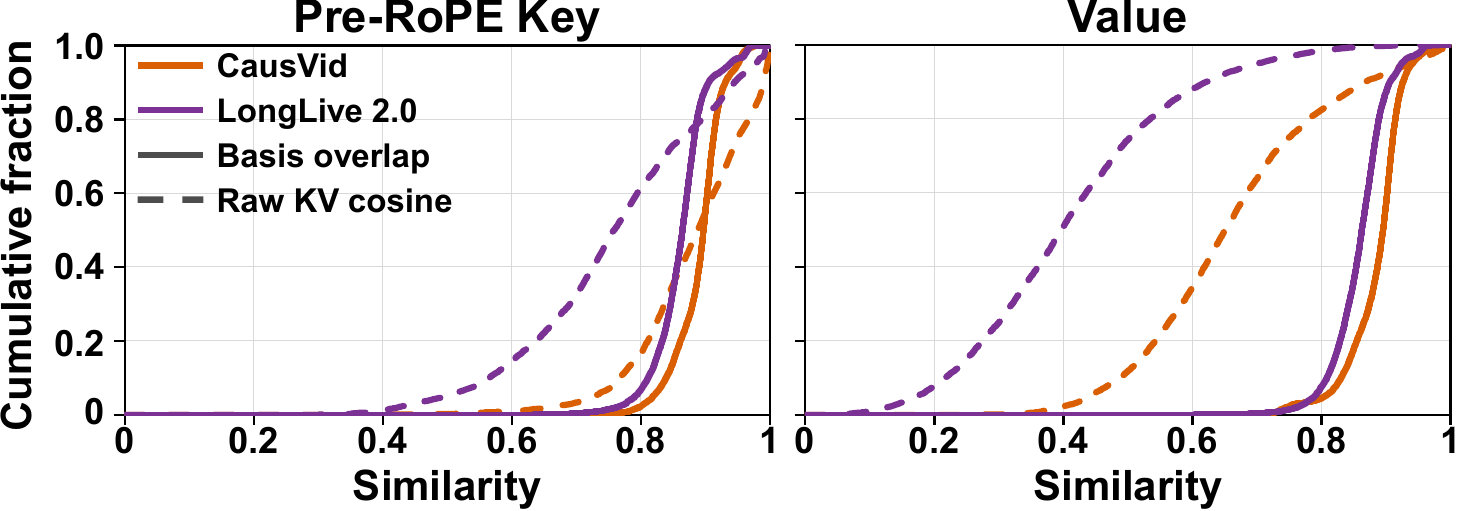}
\vspace{-9mm}
\caption{Subspace overlap and raw KV overlap of adjacent chunks under the same prompt for Key and Value at SVD rank 256.}
\vspace{-4mm}
\label{fig:basisoverlap}
\end{figure}

\vspace{-2mm}
\subsection{Shared Subspaces Across Consecutive Chunks}
\label{ss:motivation_sharingbasis}
\vspace{-1mm}
\begin{observationbox}{Observation Three:}
Adjacent chunks frequently share substantial channel structure, creating an opportunity to reuse bases.
\end{observationbox}

\vspace{-1mm}

Compressing each chunk independently would create a separate, heavy channel basis for every KV matrix. 
In practice, however, neighboring chunks often depict a continuing scene, suggesting that their KV representations may share structure across chunks.
We investigate whether this shared structure persists in the channel subspaces, even when the raw KV tensors change.


Figure~\ref{fig:basisoverlap} shows the results\footnote{Explanation of similarity metric can be found in Appendix \ref{s:c-s_similarity_across_chunks}.}. We observed that the raw KV tensors can vary substantially across consecutive chunks. For example, consecutive LongLive 2.0 chunks have mean cosine similarities of only 0.76 for the Key and 0.41 for the Value. Despite these changes, their channel subspaces remain highly similar.
At rank $r=256$, adjacent chunks generated under the same prompt have a mean channel-subspace overlap of approximately 0.89 in CausVid and 0.86 in LongLive 2.0 for both Key and Value, with 90\% of consecutive pairs exceeding 0.80.
These results indicate substantial shared channel structure across neighboring chunks despite changes in their token-level representations.

This shared structure creates an opportunity to reuse a common basis across chunks while storing separate coefficients to capture each chunk's distinct content.
However, harvesting it encounters two challenges. First, computing subspace overlap requires fitting the basis for the new chunk, which conflicts with the goal of avoiding repeated basis computation. Second, even a high overlap does not ensure that the reused basis retains enough energy from the new KV matrix. For example, two adjacent-chunk Key comparisons in the same CausVid rollout yield high overlaps of 0.951 and 0.930.
Yet, under the same storage budget as a freshly fitted basis, the preceding chunk's basis retains 98.2\% of the new matrix's energy in one case but only 96.6\% in the other, meeting the 98\% target for only the first.
\vspace{-2mm}
\section{System Design}
\label{s:design}

We present \systemname, a historical KV cache compression framework for multi-round, long video generation. It compresses the KV states of completed chunks for future reuse and restores the required historical context when a user resumes, rewinds, or branches a generation session. Figure~\ref{fig:overview} illustrates the compression and restoration workflows.

\vspace{-2mm}
\subsection{Intra-Chunk KV Compression}
\label{ss:design_intra_chunk}

\begin{figure}[t]
\centering
\includegraphics[width=\columnwidth]{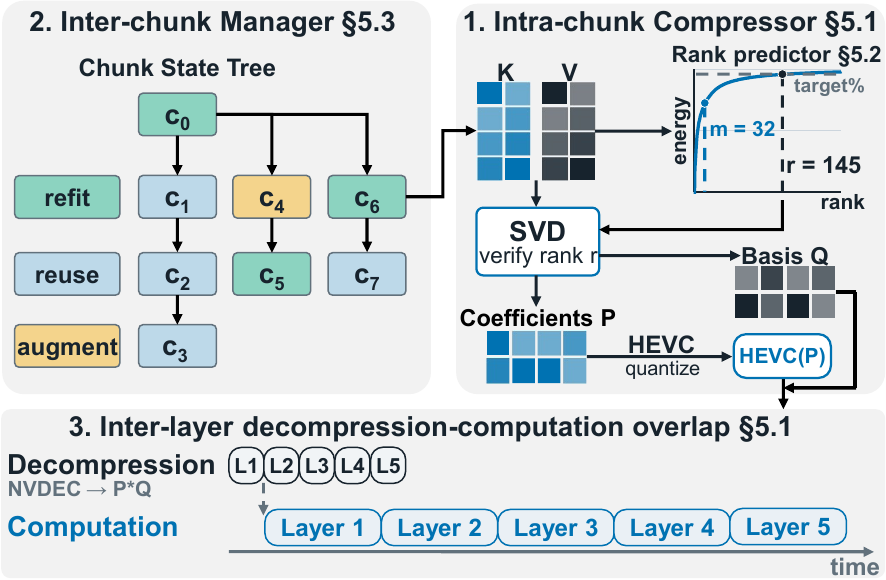}
\vspace{-8mm}
\caption{Overview of \systemname. 
}
\vspace{-5mm}
\label{fig:overview}
\end{figure}

\systemname compresses each chunk's KV through two stages: low-rank factorization across channels, followed by video coding over spatial and temporal positions. Figure~\ref{fig:overview} (part 1) illustrates the whole compression process and its reconstruction, overlapping with ongoing video generation.

\vspace{-0.5mm}
\noindent\textbf{$\bullet$ Step One: low-rank factorization.}
For each transformer layer, \systemname captures the chunk's Key before RoPE and its Value, and compresses them separately.
We represent each tensor as a matrix $A\in\mathbb{R}^{N\times C}$, where $N=Fhw$ is the number of tokens across $F$ frames, each with an $h\times w$ token grid. $C$ is the channel dimension.
\systemname approximates $A$ as:
\vspace{-1mm}
\begin{equation}\label{eq:svd}
A \approx P\cdot Q \quad
P=A\cdot Q^\top    
\end{equation}
\vspace{-1mm}
where $Q\in\mathbb{R}^{r\times C}$ contains $r$ orthonormal channel directions and $P\in\mathbb{R}^{N\times r}$ contains the corresponding coefficients for each token.
The rank $r$ determines the storage-approximation tradeoff: the coefficient matrix $P$ and the basis matrix $Q$ together contain $r(N+C)$ values, compared with $NC$ in the original matrix.
For a new factorization, \systemname fits a basis that approximates the leading right singular subspace of $A$, selecting its rank described in \S\ref{ss:design_adaptive_rank}.
The same representation also supports bases reused across chunks, as explained in \S\ref{ss:design_inter_chunk}.
\vspace{-0.5mm}

\noindent\textbf{$\bullet$ Step Two: video coding of coefficients $P$.}
Recall that in Eq.~\ref{eq:svd}, each column of $P$ contains one coefficient per token.
\systemname reshapes this column into an $F\times h\times w$ field using the tokens' original spatial and temporal positions.
It then quantizes the coefficient fields into codec-compatible samples and compresses them with HEVC.
The compressed representation consists of the encoded coefficients, the channel basis, and the metadata needed to reverse the layout and transformations. An explanation of why SVD first is in \S\ref{app:compression_pipeline_order}.

This factorization-first-then-video-coding order allows the two stages to exploit complementary redundancy: factorization reduces the number of fields to encode from $C$ original channels to $r$ coefficient channels, while video coding then exploits spatial and temporal correlations within these fields to further reduce storage.
Thus, the codec operates on the numerical coefficients produced by factorization, while the basis is retained separately for reconstruction.


\noindent\textbf{Running and restoration.}
In \systemname, the basis fitting and coefficient projection run on the GPU, while encoding runs on NVENC if available (runs on a CPU worker pool otherwise).
Note that either encoding method won't compete with ongoing video denoising for GPU's computing resources.
During restoration, NVDEC first decodes the coefficients, which are dequantized and rearranged into a reconstructed coefficient matrix $\widehat{P}$.
\systemname then computes $\widehat{A}=\widehat{P}Q$, reapplies RoPE to the reconstructed Key at its original token positions, and writes Key and Value into the GPU's memory.
To overlap decoding with generation, \systemname prepares the next layer's coefficients while the current layer computes, as shown in Figure~\ref{fig:overview} (part 3).


\vspace{-3mm}
\subsection{Online Rank Selection}
\label{ss:design_adaptive_rank}



\begin{figure}[t]
\centering
\includegraphics[width=\columnwidth]{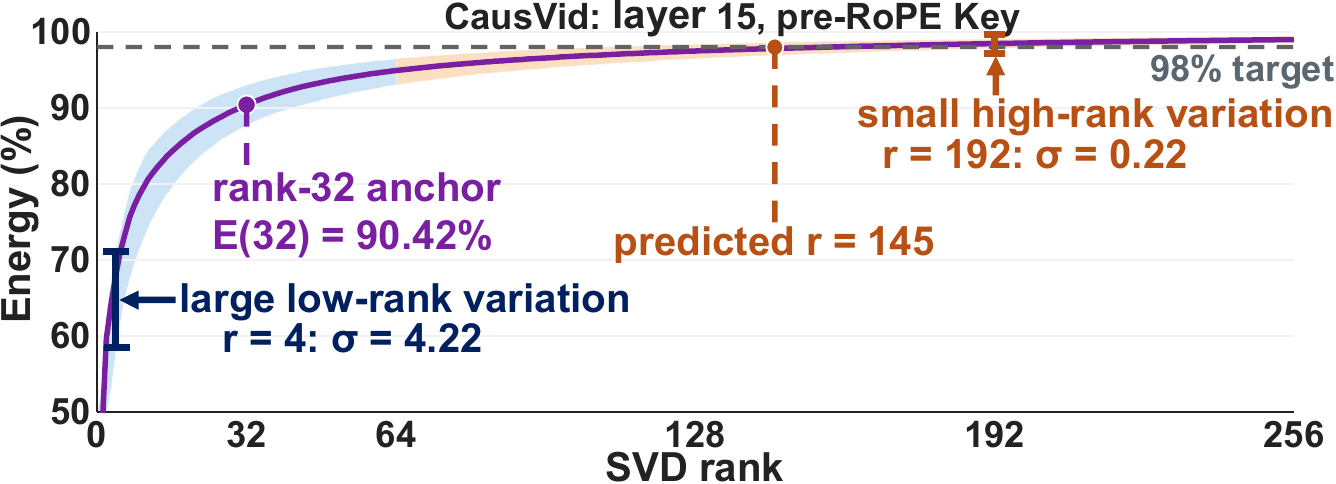}
\vspace{-9mm}
\caption{Rank-energy curve for various prompts of CausVid's layer 15. The energy variance decreases as the rank increases. Each prompt's energy curve can be found in Appendix~\ref{app:full_energy_curv}.}
\vspace{-5mm}
\label{fig:decay_pattern}
\end{figure}

\systemname estimates the rank for each layer's KV matrix individually in three steps.
It first computes a low-rank SVD of the targeting matrix $A$ and then uses the captured energy to predict the rank needed to meet the retained-energy target. 
After that, \systemname verifies the energy captured with the predicted rank and updates the predictor.
This is motivated by the observation in Figure~\ref{fig:decay_pattern}: within each layer, prompts can differ substantially in the energy captured at small ranks (the blue shaded region), but their residual energy decays at similar relative rates in the spectral tails (the orange shaded region). Accordingly, the low-rank decomposition provides us a good starting point (step one), from which the learned decay function predicts the rank required to reach the target (step two), and verifies the energy captured (step three).

To better understand this idea, consider the highlighted curve in Figure~\ref{fig:decay_pattern}. A rank-32 SVD captures 90.42\% of $A$'s energy. Using this measurement and the layer's learned decay function, \systemname predicts that a rank of 145 will retain 98\% of the energy,
without computing the full singular spectrum of $A$.
We elaborate this algorithm below.



\vspace{-1mm}
\noindent\textbf{$\bullet$ Step One: measure the captured energy.}
For a KV matrix $A\in\mathbb{R}^{N\times C}$,
\systemname constructs a basis
$S_m\in\mathbb{R}^{C\times m}$ with a low-rank of $m=32$, and then measures the captured-energy fraction $\hat E_m$ as follows:
\vspace{-1.8mm}
\begin{equation}
  \hat E_m
  = \frac{\|AS_m\|_F^2}{\|A\|_F^2}.
  \label{eq:rank-anchor-energy}
\vspace{-1.5mm}
\end{equation}
Given the basis, this measurement costs $O(NCm)$ ($m\ll C$), significantly lower than the full matrix decomposition of $A$ with $O(NC^2+C^3)$ overhead.
If $\hat E_m\ge E_t$ (the targeting energy ratio), then the basis already meets the target energy.
Otherwise, \systemname uses the measured residual energy, $1-\hat E_m$, to predict the required rank as follows.



\vspace{-1mm}
\noindent\textbf{$\bullet$ Step Two: predict the required rank.}
We approximate the residual-energy decay with a power law function and estimate the required rank using the following equation\footnote{The derivation of this equation can be found in Appendix \ref{ss:d_predictor}.}:
\vspace{-1.8mm}
\begin{equation}
  \ln\hat r
  =a_{\ell, t}
  \!+\!b_{\ell, t}\ln\!\frac{1-\hat E_m}{1-E_t}
  \label{eq:rank-online-line}
\vspace{-1.5mm}
\end{equation}
where $b_{\ell, t}>0$ describes the decay pattern for layer $\ell$ and energy target $t$. The ratio here measures how much the residual energy must decrease to meet the target. We learn $a_{\ell, t}$ and $b_{\ell, t}$ for each layer and KV type online to account for their decay patterns and approximation error in the measured basis. Both two coefficients are updated online through least square optimization, as explained below. 

\vspace{-1mm}
\noindent\textbf{$\bullet$ Step Three: energy verification and coefficient update.}
\begin{wrapfigure}{r}{0.25\textwidth}     
\vspace{-4mm}
  \centering
  \includegraphics[width=1\linewidth]{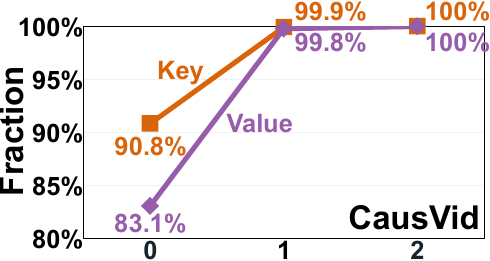}
  \vspace{-8mm}
  \caption{Extra rank adjustments required after rank prediction.}
  \label{fig:rank_predictor_cdf}
  \vspace{-6mm}             
\end{wrapfigure}
If captured energy after step 2 does not meet the target energy, \systemname adds an extra bucket of directions (32) and repeats the check until meeting the target. 
As shown in Figure~\ref{fig:rank_predictor_cdf}, more than 90\% of Keys and 83\% of Values of CausVid don't need extra rank adjustment, while one extra adjustment can cover more than 99.8\% of Keys and Values. The maximum rank adjustment is 2.


We then use these data to update the predictor online. Specifically, we use the unrounded exact rank value, $r_{\mathrm{exact}}$, as the label, so bucket rounding doesn't inflate the learned rank. For fits requiring more than the anchor rank $m$, let
$x=\ln\big((1-\hat E_m)/(1-E_t)\big)$ and $\mathbf z=(1,x)^\top$. 
Each layer maintains its online regression state, 
regularization $\mathbf M$ initialized with identity matrix $I_2$ and coefficients initialized with $\mathbf v=\mathbf M(\ln m,1)^\top$.
After verification, we update coefficients $(a_{\ell t},b_{\ell t})$ using least square optimization as follows: 
\vspace{-1mm}
\begin{equation}
  \begin{aligned}
    & \mathbf M=\mathbf M+\mathbf z\mathbf z^\top, \quad
    \mathbf v=\mathbf v
      +\mathbf z\ln r_{\mathrm{exact}}, \\
    &(a_{\ell t},b_{\ell t})^\top
      =\mathbf M^{-1}\mathbf v
  \end{aligned}
  \label{eq:rank-online-update}
\vspace{-1mm}
\end{equation}
This update incorporates the verified fit into subsequent predictions, with state retained across prompts.

\vspace{-2mm}
\subsection{Inter-Chunk KV Cache Management}
\label{ss:design_inter_chunk}


In \systemname, we store shared basis once and keep separate coefficients for each chunk. For each layer and KV type, a new chunk reuses its precedent chunk's basis if it meets the retained-energy target (e.g., 98\%). Otherwise, it adds directions to capture the missing energy, or fits a new basis when that requires less storage. We propose {\it chunk state tree} to record these basis references alongside the generation history, which can easily support resume, rewind, and forks.

\begin{figure}[t]
\centering
\includegraphics[width=\columnwidth]{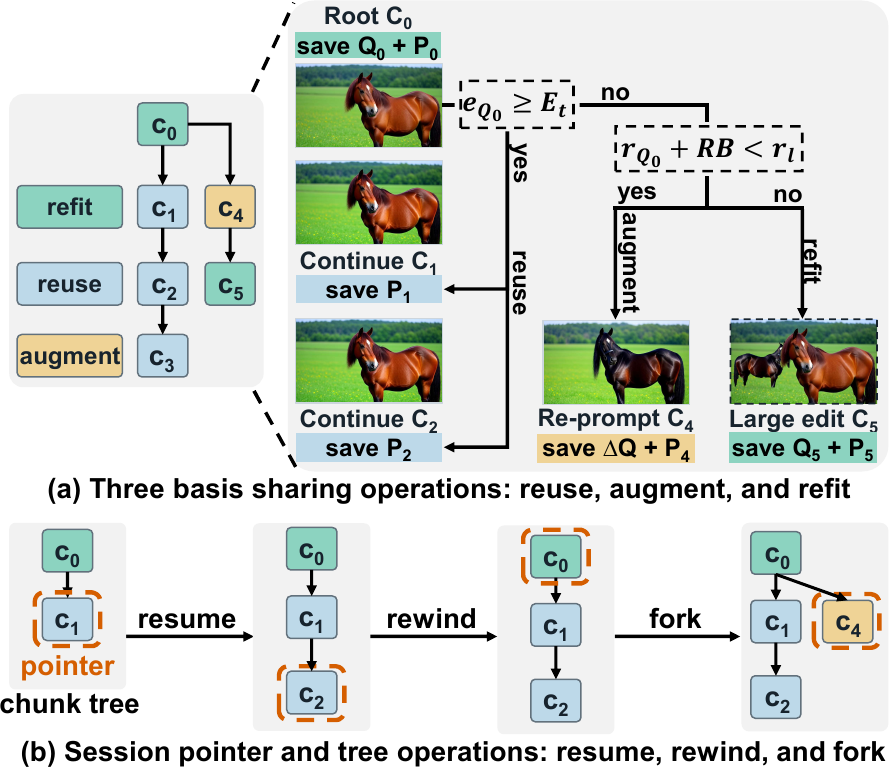}
\vspace{-8mm}
\caption{Chunk state tree in \systemname. (a) Three operations for basis sharing in consecutive chunks: reuse, augment, and refit. (b) Session pointer and user operations: resume, rewind, and fork.}
\vspace{-7mm}
\label{fig:cache_tree}
\end{figure}

\noindent\textbf{Chunk state tree.}
As shown in Figure~\ref{fig:cache_tree}(a), each node represents a chunk and records
its HEVC-encoded coefficients, references to the basis directions it uses, and
reconstruction metadata.
Each edge connects a parent chunk to its child, so a root-to-node path, such as $c_0
\rightarrow c_1 \rightarrow c_2 \rightarrow c_3$, represents a consecutive video
generation history.

For each new chunk, \systemname first attempts to reuse its parent's basis.
If this basis captures sufficient energy to meet the target threshold, which is $e_{Q_0} \geq E_t$, the child stores its own HEVC-encoded coefficients while
referencing the existing basis.
For example, chunks $c_1$ and $c_2$ store separate coefficients $P_1$ and
$P_2$ but share the stored basis $Q_0$.

If the inherited basis falls short of the target, \systemname incrementally extends it with additional basis directions $\Delta Q$, adding \textit{RB} directions at a time (e.g., 32).
Once the extended basis meets the target, the child stores the additional directions $\Delta Q$ and coefficients for the extended basis, as illustrated by chunk $c_4$.
If the required total rank exceeds the rank used in the most recent fit $r_l$, \systemname
instead refits a new basis with the methods in
\S\ref{ss:design_intra_chunk} and \S\ref{ss:design_adaptive_rank}.
For example, the edited chunk $c_5$ differs substantially from its ancestor $c_0$, triggering a refit that produces a new basis $Q_5$.

The session maintains a pointer to its current chunk node, as shown in Figure~\ref{fig:cache_tree}(b).
The multi-round video generation then becomes tree operations: {\it resume} continues from this node; {\it rewind} selects an ancestor; and {\it fork} starts a new branch whose first chunk is a child of any existing node and leaves the parent's branch untouched. Existing nodes remain unchanged, allowing branches to share their common history without copying it. 
The detailed algorithm for determining if a node can share the basis of its parent node is in \S\ref{s:inter_chunk_basis_reuse}.

\vspace{-2.5mm}
\section{Evaluation}
\label{s:eval}

We implement \systemname in approximately 21K LoC and integrate it into SGLang-Diffusion \cite{sglang} with 2K LoC. \systemname adopts LRU to manage cache. 

\vspace{-2.3mm}
\subsection{Experimental Setup}


\noindent\textbf{Testbed.} We conduct experiments on a server with an AMD EPYC 9555 64-core processor, eight RTX PRO 6000 GPUs with 96GB GDDR7 each, and 512 GB host memory.
\\
\noindent\textbf{Models.} We evaluate five autoregressive DiT models with three model sizes.
CausVid~\cite{causvid}, RAVEN~\cite{raven}, and LongLive~1.0~\cite{longlive} have  1.3B parameters.
LongLive~2.0~\cite{longlive2.0} has 5B parameters, and LingBot~\cite{lingbot} has 13.6B parameters.
We preserve each model's native generation granularity, attention visibility, positional encoding, and conditioning semantics.
\\
\noindent\textbf{Datasets.} Since no public multi-round video generation dataset is available for evaluation, we build the \emph{Synthetic-500} dataset by extending MemFlow's
interactive multi-prompt benchmark~\cite{ji2025memflow}
to 500 narratives with 10,000 main-path and 5,000 alternative
prompts. An additional 1,328 prompts support branches visited
during evaluation. Appendix~\ref{app:corpus} details the corpus.
\\
\noindent\textbf{Workloads.}
We generate requests and set up different workloads for evaluation.
We characterize workloads by two parameters:
\emph{request rate}: the aggregate number of submitted prompts per second; and
\emph{rewind frequency}: the probability of selecting a non-latest checkpoint.
Rewind generates a new branch to a random historical checkpoint. Other requests extend the latest checkpoint.
We fix rewind frequency to 0.2 in \S\ref{ss:eval_main_results}.
In \S\ref{ss:workload_sensitivity}, we perform the sensitivity study over rewind frequencies and measure \systemname's speedup.
\\
\noindent\textbf{Metrics.}
For storage, we report the compression ratio: the size of the uncompressed BF16 KV divided by the bytes \systemname stores for it.
For video quality, we measure VBench scores\cite{vbench}, a gold-standard evaluation metric suite that quantifies video generation quality, and SSIM to measure frame-wise quality.
\\
\noindent\textbf{Baselines.}
We compare \systemname with two baselines: \emph{Recompute} and \emph{Offload}.
\emph{Recompute} keeps no history KV outside the active GPU cache.
When a session resumes or rewinds, it rebuilds the historical KV by replaying the forward passes for video latents.
\emph{Offload} keeps the uncompressed KV of generated chunks in GPU memory and host memory, evicting the least recently used chunks when storage is full.
When a request needs evicted KV, it recomputes the missed KV from video latents. 

\subsection{End-to-End Performance Comparison}
\label{ss:eval_main_results}

\begin{figure}[t]
\centering
\includegraphics[width=\columnwidth]{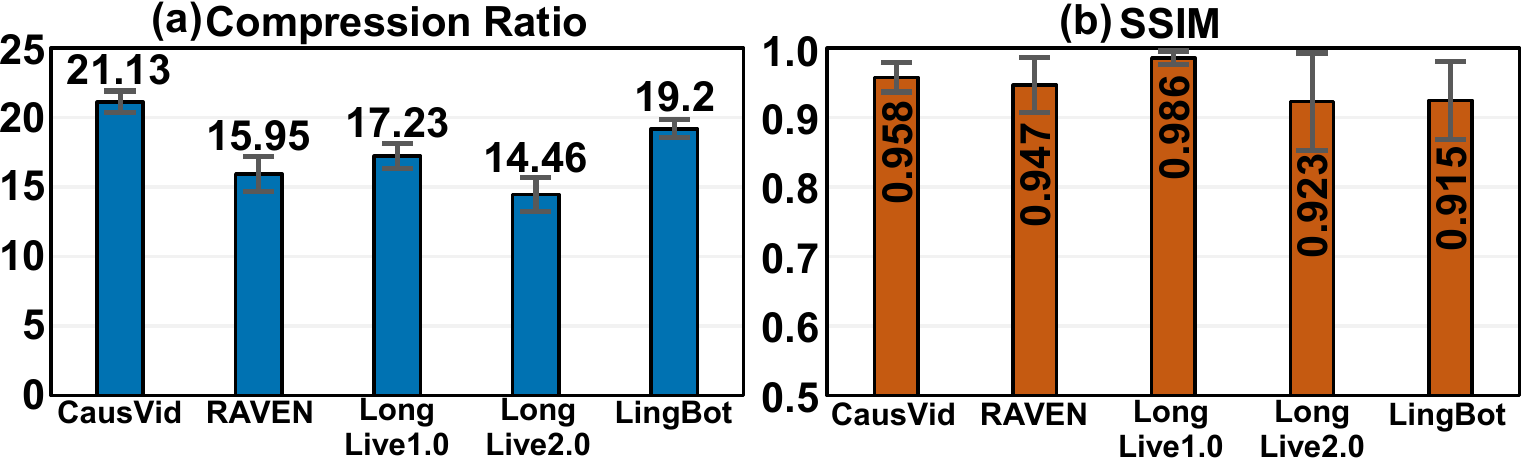}
\vspace{-9mm}
\caption{Mean and standard deviation of compression ratio and SSIM for models under \systemname. Higher is better.}
\label{fig:video_ratio_ssim}
\vspace{-3mm}
\end{figure}

\textbf{Compression Ratio.} Figure~\ref{fig:video_ratio_ssim}(a) reports the mean KV compression ratio and standard deviation for all five models.
\systemname achieves mean compression ratios ranging from 14.46$\times$ for LongLive~2.0 to 21.13$\times$ for CausVid. For RAVEN, LongLive~1.0, and LingBot, the corresponding ratios are 15.95$\times$, 17.23$\times$, and 19.20$\times$, respectively. The standard deviations are small relative to the means, indicating limited variation across the evaluated samples within each model. These reductions allow \systemname to retain substantially more historical KV under a fixed storage budget than the
uncompressed \emph{Offload} baseline, potentially reducing
evictions and subsequent recomputation under memory pressure.

\begin{figure}[t]
\centering
\includegraphics[width=\columnwidth]{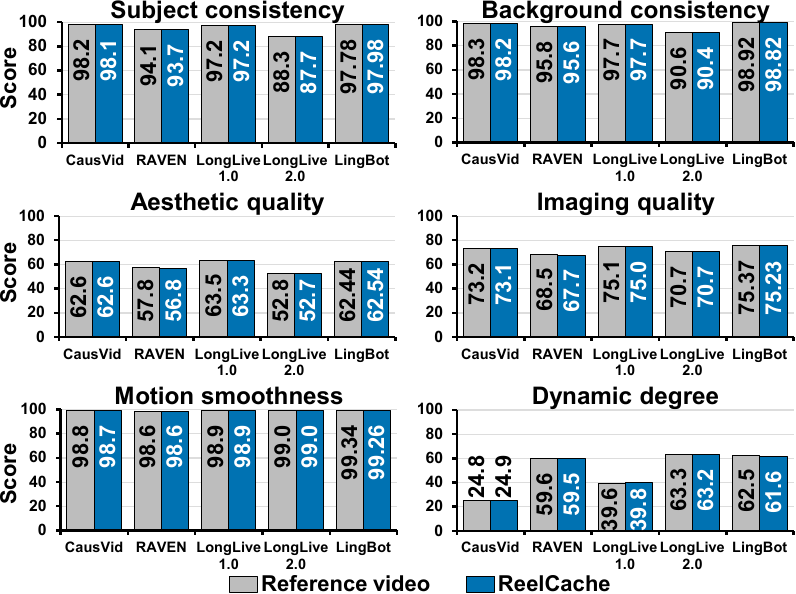}
\vspace{-8mm}
\caption{Six-dimensional VBench scores between reference videos and videos generated by \systemname.}
\label{fig:video_vbench}
\vspace{-6mm}
\end{figure}

\textbf{Video Quality.} We next examine whether this KV compression comes at the cost of video quality. Figure~\ref{fig:video_vbench} compares generated videos against matched references produced without KV compression using six representative VBench dimensions~\cite{vbench}. 

Across all five models, \systemname's scores closely track the uncompressed references on every dimension. For example, motion smoothness changes by at most 0.1 points, while subject and background consistency remain within 0.6 points of the references. Aesthetic quality, imaging quality, and dynamic degree also remain close, with the largest reduction across all six dimensions being 0.97 points on the 0--100 scale. These results show that \systemname substantially reduces
KV storage while largely preserving video quality. We show a few frame comparisons in Appendix~\ref{app:video_frame}.

 
We also assess frame-wise fidelity using SSIM against
the corresponding reference frames.
Figure~\ref{fig:video_ratio_ssim}(b) reports the mean SSIM and
standard deviation for each model.
All five models achieve mean SSIM values of at least 0.915,
ranging from 0.915 for LingBot to 0.986 for LongLive~1.0.
These results indicate that the generated videos retain high
structural similarity to their uncompressed references.
 



\begin{figure*}[t]
\centering
\includegraphics[width=\linewidth]{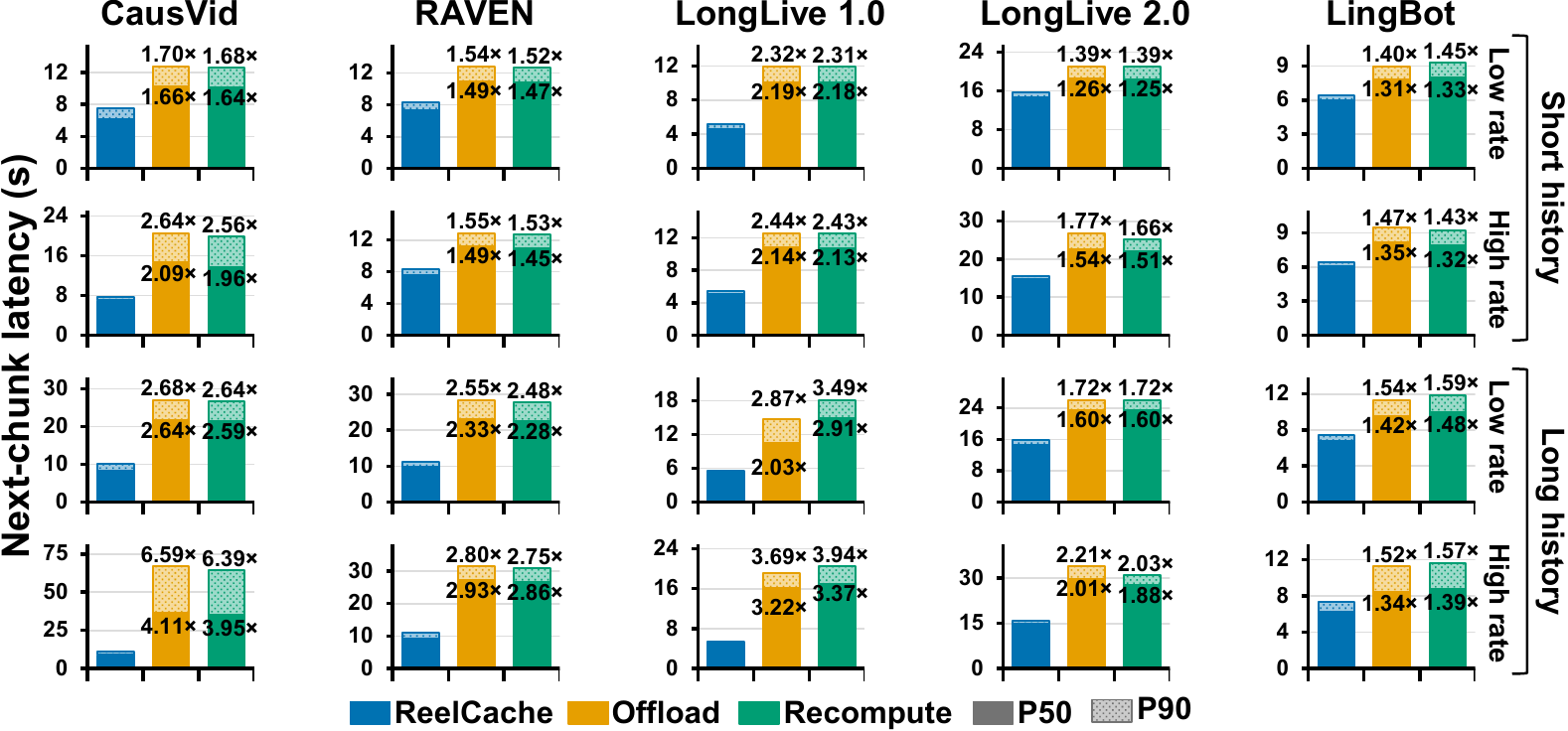}
\vspace{-8mm}
\caption{P50 and P90 next-chunk latency of \systemname compared to baselines across models and workloads.}
\vspace{-2mm}
\label{fig:next_chunk_latency}
\end{figure*}

\textbf{Next-Chunk Latency.} We compare the P50 and P90 next-chunk latencies under four workload configurations combining short/long histories (20/40 chunks) with low/high request loads. As shown in Figure~\ref{fig:next_chunk_latency}, across all 20 model-workload combinations, \systemname achieves mean P50 speedups of 2.01$\times$ over \emph{Offload} and 2.03$\times$ over \emph{Recompute}. The corresponding P90 speedups are 2.29$\times$ and 2.36$\times$.

 
Specifically, when the request load is low, \systemname remains beneficial, providing mean P50 speedups of 1.83$\times$ over both baselines. Its average tail-latency advantage is more pronounced under long histories and high request loads: the mean P90 speedup over \emph{Offload} increases from 1.65$\times$ in the short-history, low-load configuration to 3.36$\times$, while the corresponding speedup over \emph{Recompute}
increases from 1.67$\times$ to 3.34$\times$.
 
This benefit is because within a fixed memory budget, \systemname can retain more
compressed KVs than \emph{Offload}, reducing
capacity-induced misses and their reconstruction costs.
\emph{Offload} transfers retained raw KV back to the GPU and
rebuilds required KV from latents when it has been evicted.
\emph{Recompute} instead retains latents rather than a
historical KV archive and reconstructs the required context
through model forwards.
\systemname restores compressed checkpoints on demand, avoiding the expensive recomputation costs and reducing waiting times for queued requests.

\subsection{System Efficiency}
We then evaluate system efficiency by measuring the time \systemname spends on compressing and decompressing KV. We also evaluate how much compression and decompression operations slow down next-chunk generation, respectively. 

\begin{figure}[t]
\centering
\includegraphics[width=\columnwidth]{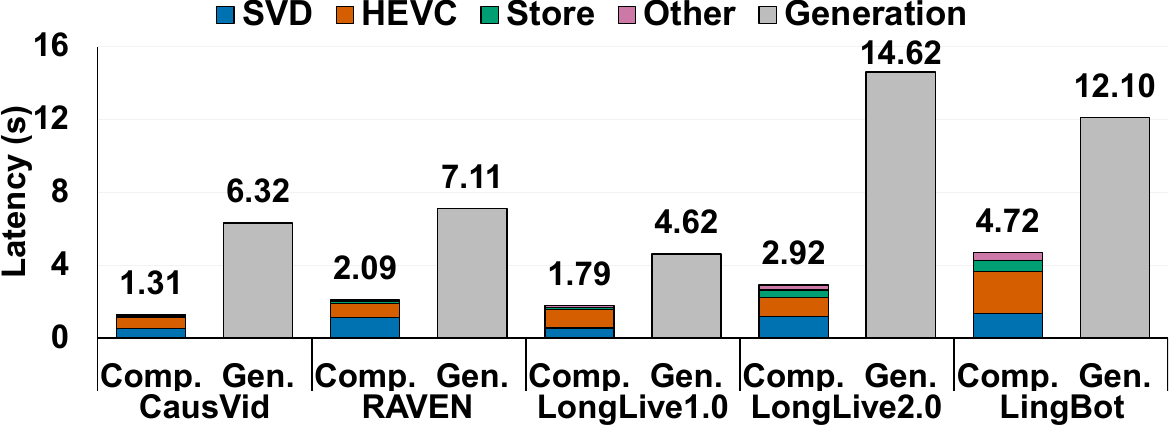}
\vspace{-8mm}
\caption{Compression time breakdown in \systemname and model generation time.}
\vspace{-2mm}
\label{fig:system_overhead}
\end{figure}

\noindent\textbf{Compression.} Figure~\ref{fig:system_overhead} lists the time breakdown of the chunk compression time and the comparison with the chunk generation time.
Compressing a chunk takes 1.31-4.72s, 0.2-0.4$\times$ the time to generate the KV.
SVD (rank selection, fitting, projection, and verification) and HEVC encoding are the two most expensive operations, together accounting for 88\% of the compression time on average. Note that \systemname keeps this work off the critical path: it schedules SVD on a dedicated CUDA stream, and a chunk's SVD is deferred while a video chunk is generating and runs between chunks, unless uncompressed chunks will be offloaded to host memory by the cache policy.
HEVC encoding runs on the CPU or NVENC engine, so it proceeds during generation without occupying the streaming multiprocessors.
Interference therefore arises only when a request arrives while an SVD is still running, and the request then shares the GPU with the remainder of that SVD. As listed in Table~\ref{tab:overhead_slowdown}, compression slows an ongoing video generation job running on the same GPU by at most 0.8\%.

\noindent\textbf{Decompression.} \systemname overlaps decompression with video generation layer by layer: NVDEC decodes the coefficients of the next layer while the current layer computes, and reconstructing each K or V matrix requires only one matrix multiplication of its coefficients and basis. 
Generation stalls only when the first forward pass reaches a layer whose KV is not yet restored.
Because full-attention models restore every history chunk,  decompression slows video generation by 3.2\% for CausVid, 4.2\% for RAVEN, and 3.7\% for LingBot.
Sliding-window attention-based LongLive~1.0 and LongLive~2.0 restore only the limited chunks in their attention window, regardless of history length. Hence decompression slows their generation by merely $<$0.1\% and 0.6\%, respectively.
These delays are far less than the latency \systemname saves over KV recomputation (\S\ref{ss:eval_main_results}).

\subsection{Ablation Study}
\label{ss:ablation}
We next quantify the contribution of each design part toward compression ratio.
\textit{Stage-1: HEVC compression} that encodes KV with HEVC. 
\textit{Stage-2: SVD compression} that decomposes KV with SVD. 
\textit{Stage-3: SVD+HEVC compression pipeline} that decomposes KV with SVD first, then encodes SVD coefficients with HEVC.
\textit{Stage-4: Adaptive rank} that dynamically adjusts SVD rank across layers during compression. 
We designate \textit{Stage-5} as the standard \systemname configuration, representing the fully implemented system on our testbed. For the fixed-rank SVD configurations in Stages 2 and 3, we use $r_K=C/6$ and $r_V=C/4$, where $C$ denotes the channel dimension.

\begin{figure}[t]
\centering
\includegraphics[width=\columnwidth]{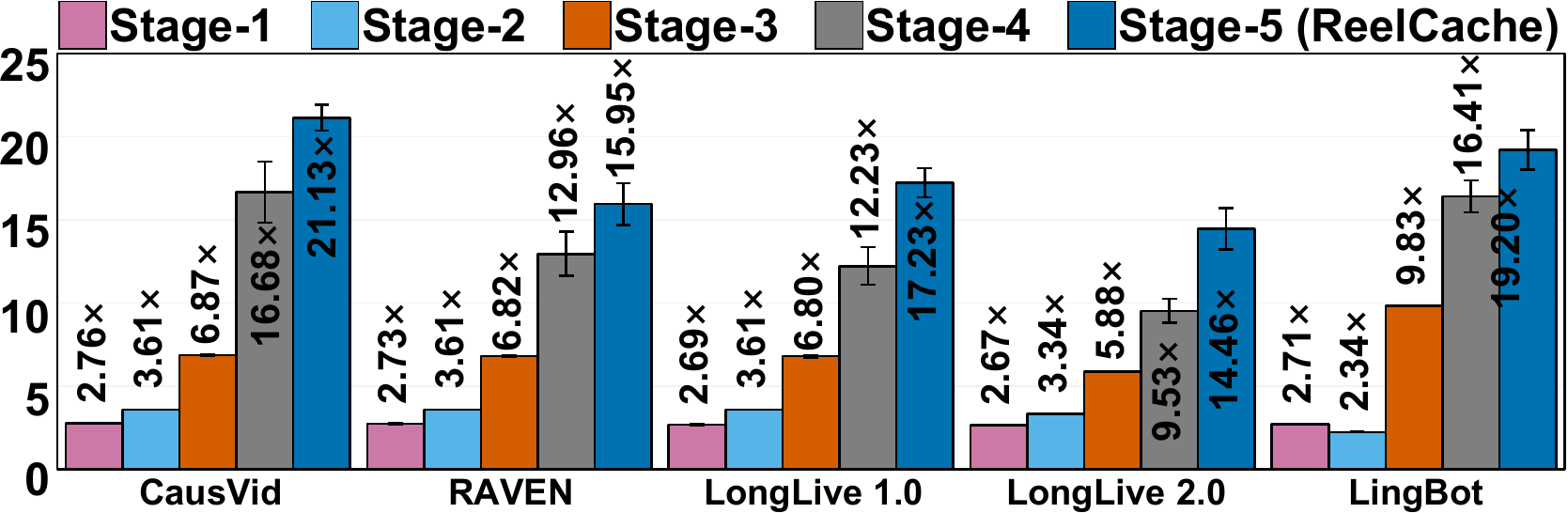}
\vspace{-8mm}
\caption{Breakdown of average compression ratio across incremental stages, where Stage-5 is \systemname.}
\vspace{-5mm}
\label{fig:ablation_compression_ratio}
\end{figure}

Figure~\ref{fig:ablation_compression_ratio} breaks down the compression ratio of each stage.
Stage-1 (HEVC only) and Stage-2 (SVD only) compress the least, by an average of 2.71$\times$ and 3.30$\times$ across models.
Each applies a single transform/encoding and therefore removes only one kind of redundancy (intra-channel for HEVC, inter-channel for SVD), leaving the other on the table.
Combining the two in Stage-3 nearly doubles the ratio to 7.24$\times$ (2.19$\times$ over Stage-2).
Stage-3, however, applies one fixed rank to every prompt and layer, so many layers keep far more rank than they need to reach the energy target (\S\ref{ss:motivation_rank}), which caps the achievable compression.
Enabling the adaptive rank for layers in Stage-4 removes this excess rank and raises the mean ratio by 1.87$\times$, from 7.24$\times$ to 13.56$\times$.
Finally, Stage-5, the default \systemname{} configuration, shares bases across consecutive chunks.
This basis sharing further reduces storage by 22.9\%, raising the mean compression ratio to 17.59$\times$.

We further examine \systemname's performance to workload sensitivity and put the results in Appendix~\ref{ss:workload_sensitivity}.

\begin{table}[t]
\centering
\small
\vspace{-4mm}
\caption{Slowdown of next-chunk generation caused by \systemname's compression and decompression.}
\label{tab:overhead_slowdown}
\begin{tabular}{llrr}
\toprule
Model        & Attention      & Compress & Decompress \\
\midrule
CausVid      & Full           & 0.2\%     & 3.2\% \\
RAVEN        & Full           & 0.8\%     & 4.2\% \\
LingBot      & Full           & 0.1\%     & 3.7\% \\
LongLive 1.0 & Sliding window & $<$0.1\%  & $<$0.1\% \\
LongLive 2.0 & Sliding window & $<$0.1\%  & 0.6\% \\
\bottomrule
\end{tabular}
\vspace{-4mm}
\end{table}

\section{Conclusion}
We have presented the design, implementation, and evaluation of \systemname, an efficient KV cache compression and management system for multi-round video generation. Within each chunk, it leverages SVD and video codec to remove both inter- and intra-channel redundancy.
Besides, it employs an online adaptive rank determination to reduce storage of bases and coefficients without degrading video quality.
Across chunks, \systemname maintains a chunk state tree and dynamically shares bases across consecutive chunks for less SVD basis storage.  It also enables basis augmentation and basis refit when necessary to maintain the video quality. \systemname reduces KV storage by 14.46$\times$-21.13$\times$ with less than 1\% drop in VBench metrics.

\bibliography{reference}
\bibliographystyle{mlsys2025}

\section*{AI Usage Disclosure}
\label{s:ai_disclosure}
We used generative AI tools in the following parts of this work.

\noindent\textbf{Code generation.}
We used Fabel 5.1 as a coding assistant to implement parts of \systemname.
All AI-assisted code was reviewed, tested, and validated by the authors.

\noindent\textbf{Data generation.}
We used GPT-6 Astra to author 400 of the 500 narratives in the \emph{Synthetic-500} evaluation corpus (Appendix~\ref{app:corpus}) that extend each script in MemFlow~\cite{ji2025memflow} to 20 consecutive
prompts.
We also leverage GPT-6 Astra to write 5,000 alternative prompts and 1,328 additional prompts to support user rewind and fork operations in multi-round video generation workloads.
These prompts serve only as workload inputs.

\noindent\textbf{Writing and references.}
We used GPT-6 Astra to rephrase and polish the text for clarity and readability.

Generative AI was not used for research ideation, the design of \systemname, experiment design, data analysis, or generating any of the results, claims, or findings reported in this paper.
We have reviewed all AI-assisted work and take full responsibility for the content of this submission.

\newpage
\appendix

\section{Why SVD First Then Video Codec}
\label{app:compression_pipeline_order}

Leveraging both SVD and a video codec (e.g., HEVC) offers a high compression ratio, as shown in Figure~\ref{fig:compression_pipeline}.
\systemname employs a SVD-first-then-HEVC compression pipeline.
Reversing the order, HEVC followed by SVD, changes the object being approximated: SVD operates on an entropy-coded byte stream rather than the KV tensor. 
Its numerical approximation does not enforce the syntax required by the HEVC decoder.
All video chunk streams reconstructed after SVD failed decoding in our offline experiments.

\begin{figure}[t]
\centering
\includegraphics[width=\columnwidth]{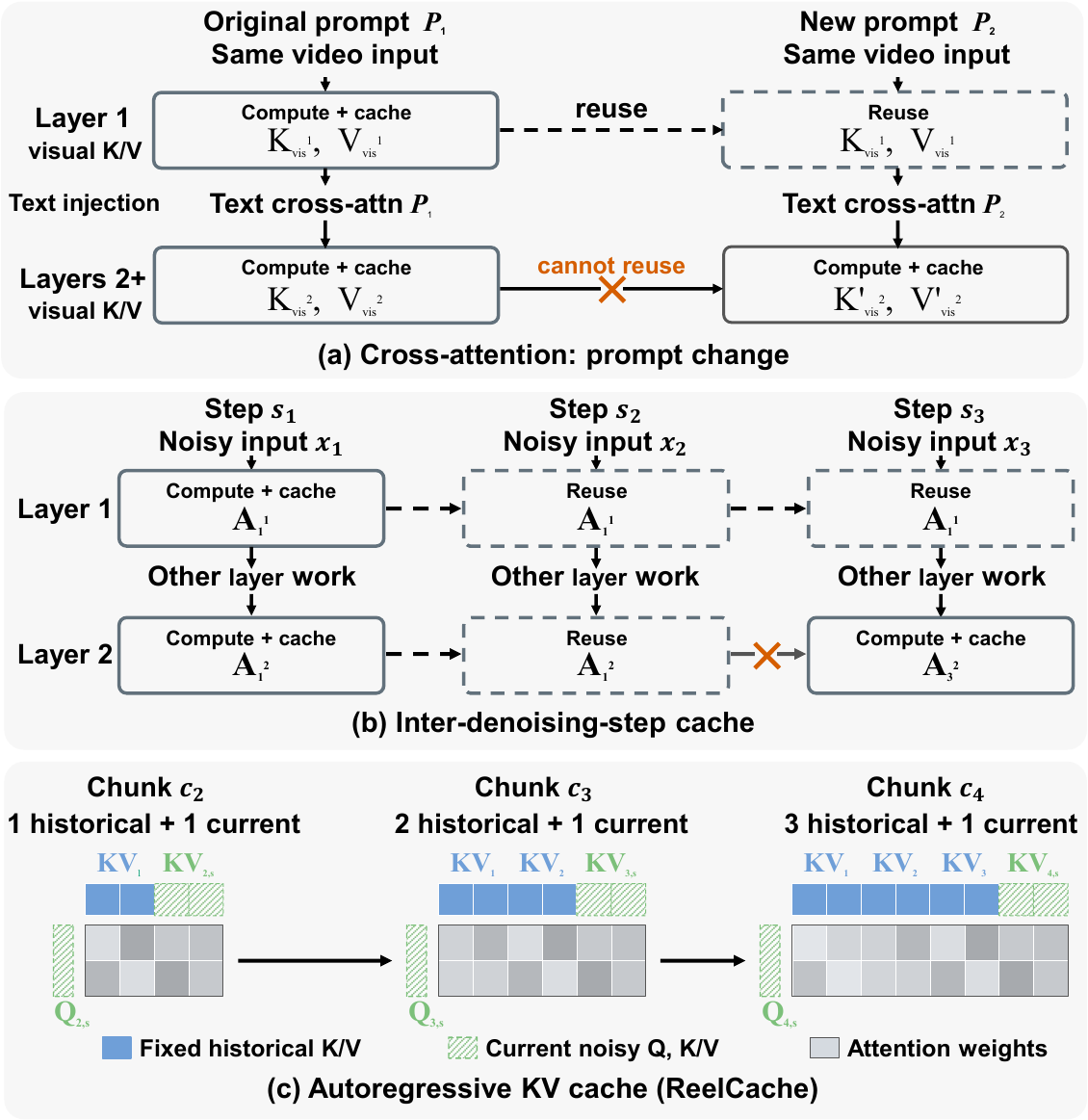}
\vspace{-6mm}
\caption{Illustration for (a) KV reuse in cross-attention, (b) intermediate activation cache during denoising, and (c) \systemname's historical KV cache in autoregressive video generation.} 
\label{fig:kv_conditioning}
\end{figure}

\section{Autoregressive Diffusion Transformers}
\label{app:cache_activations}

In \systemname, we target chunk-autoregressive video diffusion
transformers (AR-DiTs) that expose reusable, layer-wise
self-attention keys and values for completed video
context~\cite{causvid,longlive}.
These models generate video sequentially, with each new chunk
attending to the cached KV of selected preceding chunks.
This causal historical KV therefore conditions subsequent chunks
without repeatedly processing the preceding video, subject
to the model's attention window.
\systemname compresses and retains this historical state
for efficient restoration when generation resumes or branches
from an earlier point.

The historical video KV targeted here is distinct from
text-conditioning KV used in cross-attention, where visual
queries attend to keys and values derived from text
representations~\cite{crossattn_Chen_2024_CVPR,
crossattn_chen2024pixartalpha,crossattn_Photorealistic,
crossattn_Rombach_2022_CVPR}, as depicted in Figure~\ref{fig:kv_conditioning}(a).
Visual tokens that are jointly updated during
bidirectional denoising or in-place editing generally produce
step- or edit-dependent hidden states, whose KV cannot be
assumed to remain valid as an unchanged historical prefix.
Hence, these scenarios are out of \systemname's scope. 

Our objective also differs from intermediate-activation
caching~\cite{teacache,
activation_cache_lv2025fastercache,deepcache,
activation_cache_pyramid_kv,activation_cache_Wimbauer_2024_CVPR}.
These methods exploit similarity between denoising steps
to reuse attention outputs, block features, or other
intermediate results, reducing computations
within the denoising process, as shown in Figure~\ref{fig:kv_conditioning}(b).
In contrast, \systemname preserves the KV of completed
video chunks for subsequent chunk generation and interactive
continuation.
It accelerates recovery of the required historical context,
rather than skipping the denoising computations of the new
chunk, as illustrated in Figure~\ref{fig:kv_conditioning}(c).

\section{Efficient and parallel video inference}
\label{s:related_efficient_parallel_video_infer}

Two orthogonal lines make video-DiT inference practical, and \systemname coexists with both. Step-cache works~\cite{pipefusion,distrifusion,teacache,deepcache} exploit the temporal redundancy of the denoising loop and skip recomputation for nearly identical activations. Parallel-serving systems, on the other hand, distribute the model or tokens with sequence parallelism (SP), because a video DiT's 3D activations and KV cache might exceed one GPU's capacity. For instance, Ulysses~\cite{ulysses} shards the token sequence where each GPU owns whole attention heads (\emph{head/channel} split); RingAttention~\cite{ringattention} keeps a contiguous token shard and rotates keys/values around a ring (\emph{token} split). USP~\cite{usp} composes both.
Inference engines~\cite{xdit,vllm-omni,sglang} integrate SP with other parallelism (e.g., pipeline), and recent work~\cite{latentparallelism} proposes latent parallelism for video diffusion to cut the heavy cross-GPU activation traffic.
SP fixes the cache layout shared by head/token that any KV compressor must accept before compression runs.

\begin{figure}[t]
\centering
\includegraphics[width=\columnwidth]{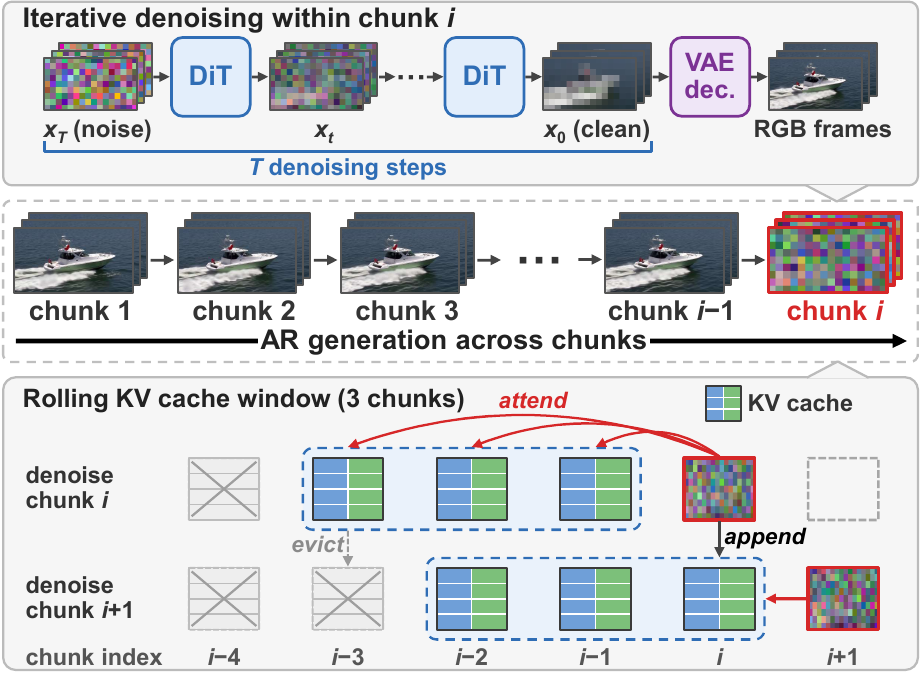}
\vspace{-6mm}
\caption{Autoregressive video generation workflow. Middle: the model generates a video chunk by chunk, each chunk conditioned on the preceding ones. Top: within chunk $i$, the DiT iteratively denoises noisy latents $x_T$ into clean latents $x_0$ over $T$ steps, and the VAE decodes $x_0$ into RGB frames. Bottom: the KV of the most recent chunks is kept in a rolling cache window (three chunks shown). Chunk $i$ attends to the cached KV. Once it is generated, its KV is appended, and the oldest KV$_{i-3}$ is evicted.}
\label{fig:background}
\vspace{-4mm}
\end{figure}


\section{Autoregressive Diffusion Video Generation Primer}
\label{ss:avgp}

Modern autoregressive video diffusion models, such as CausVid~\cite{causvid} and LongLive~\cite{longlive,longlive2.0}, generate a video one chunk at a time, conditioning each new chunk on previously generated content.
As illustrated in Figure~\ref{fig:background} (top), generation proceeds at two timescales: {\it autoregressive generation across chunks} and {\it iterative denoising within each chunk}.
For each chunk, a diffusion transformer (DiT) starts from noisy latent tokens and progressively refines them through multiple denoising steps, conditioned on a text prompt and the preceding video context.
A variational autoencoder (VAE) then decodes the resulting clean latents into RGB frames.

To maintain temporal consistency, the DiT denoises each new chunk while attending to previously generated content.
Models such as LongLive~\cite{longlive} cache the layer-wise keys and values (KV) of preceding chunks within a rolling window, allowing subsequent chunks to reuse these representations without recomputing them, thereby reducing a significant amount of video generation latency.
As illustrated in Figure~\ref{fig:background} (bottom), this cache gets updated with generation, enabling efficient continuation while evicting older entries outside the retained context.

\section{channel-subspace similarity across chunks}
\label{s:c-s_similarity_across_chunks}

To measure channel-subspace similarity across chunks, we use normalized subspace overlap:
\[
    s(i,j) = \frac{\|Q_i Q_j^{\top}\|_F^2}{r},
\]
where $Q_i$ and $Q_j$ are row-orthonormal rank-$r$ channel bases for corresponding KV matrices in chunks $i$ and $j$, respectively.
This score is the mean squared cosine of the principal angles between the two subspaces and is invariant to the choice of orthonormal basis within each subspace.

\begin{figure}[t]
\centering
\includegraphics[width=\linewidth]{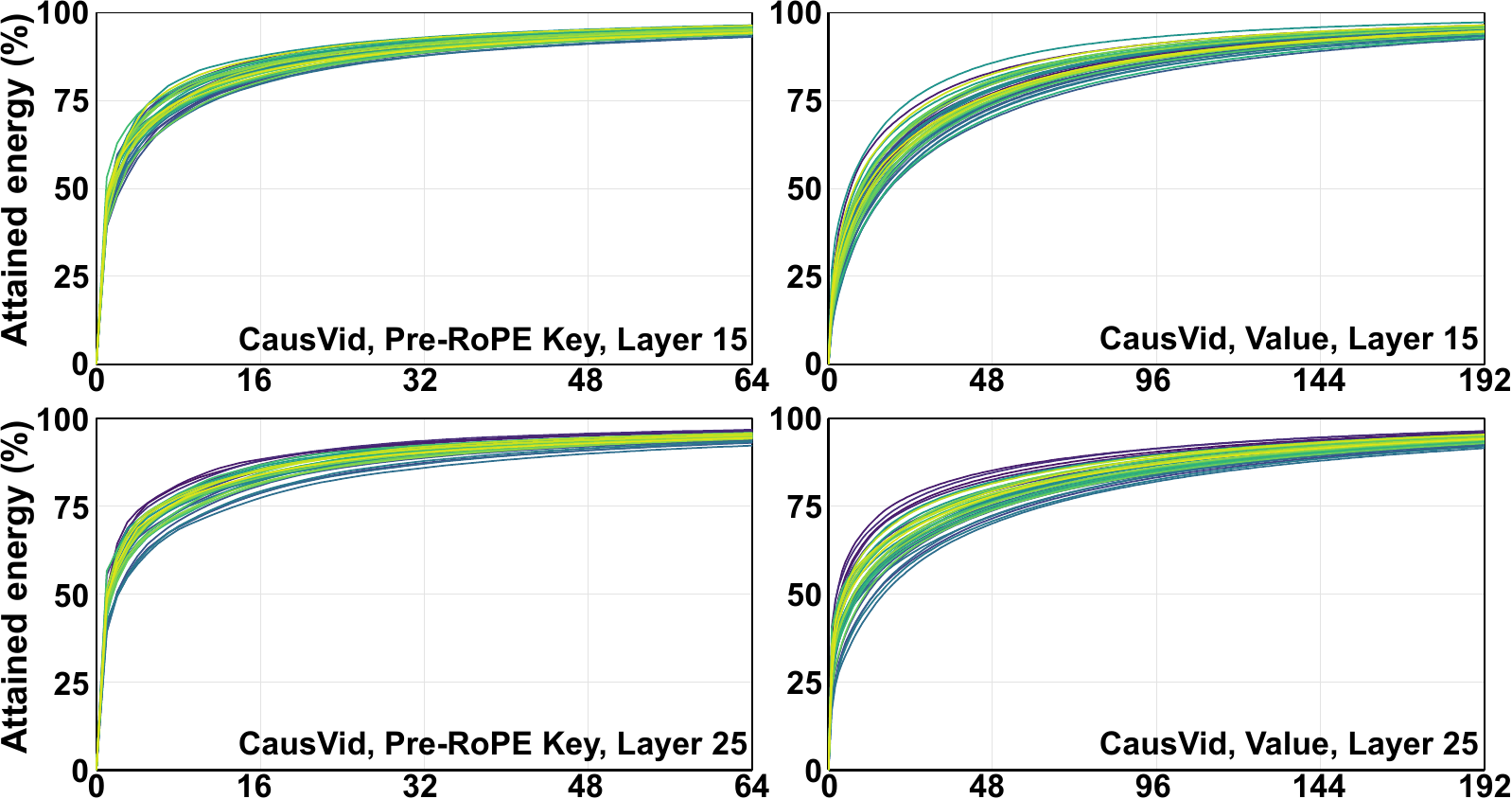}
\vspace{-8mm}
\caption{Prompts' energy curves of CausVid's layers 15 and 25.}
\label{fig:full_energy_curve}
\end{figure}

\section{Full Energy Curve Plot}
\label{app:full_energy_curv}
Figure~\ref{fig:full_energy_curve} lists energy curves of prompts for CausVid's layers 15 and 25 (pre-RoPE Key and Value). 
Within each layer, prompts can differ substantially in the energy captured at small ranks, but their residual energy decays at similar relative rates in the spectral tails.

\section{Derivation of the predictor}
\label{ss:d_predictor}
We approximate the residual-energy decay with a power law function:
\begin{equation}
  E_X(r)
  \approx 1-\big(1-E_X(m)\big)
                 (r/m)^{-\gamma_{\ell, t}},
  \qquad r\ge m
  \label{eq:rank-energy-model}
\end{equation}
where $E_X(r)$ denotes the energy fraction captured by the optimal rank-$r$ approximation, $\gamma_{\ell, t}>0$ describes the decay pattern for layer $\ell$ and energy target $t$. 
Substituting $E_X(m)$ in above equation with $\hat E_m$ measured by Eq.~\ref{eq:rank-anchor-energy} and further solving for the rank at which the captured energy reaches $E_t$ gives:
\begin{equation}
  \hat r
  \approx m\left(
    \frac{1-\hat E_m}{1-E_t}
  \right)^{1/\gamma_{\ell, t}}
  \label{eq:rank-anchor-prediction}
\end{equation}

Taking the logarithm on both sides, we have a rank estimator:
\begin{equation}
  \ln\hat r
  \approx \ln m
  \!+\!\frac{1}{\gamma_{\ell t}}
  \ln\!\frac{1-\hat E_m}{1-E_t}
  =a_{\ell, t}
  \!+\!b_{\ell, t}\ln\!\frac{1-\hat E_m}{1-E_t}
  \label{eq:rank-estimator}
\end{equation}
The ratio here measures how much the residual energy must decrease to meet the target. We learn $a_{\ell, t}$ and $b_{\ell, t}$ separately for each layer and KV type to account for their decay patterns and approximation error in the measured basis. In the online estimation, both coefficients $a_{\ell, t}$ and $b_{\ell, t}$ are updated through least-squares optimization. 

\begin{algorithm}[t]
\caption{Commit matrix $A$ of a new chunk, given the branch's active basis $Q$.}
\label{alg:commit}
\begin{algorithmic}[1]
\small
\STATE $P \leftarrow A\tilde Q^{\top}$,\quad
       $R \leftarrow A-P\tilde Q$ \algcomment{project}
\STATE $e \leftarrow 1-\|R\|_F^{2}/\|A\|_F^{2}$
       \algcomment{certify}
\IF{$e \ge E_t$}
  \STATE keep the shortest prefix of $Q$ reaching $E_t$
         \algcomment{share}
    \STATE $Q \leftarrow \tilde Q$
\ELSE
  \STATE $U \leftarrow \textsc{Fit}\big(R,\ (E_t-e)/(1-e)\big)$
         \algcomment{new directions}
  \IF{storing $[\tilde Q,U^{\top}]$ costs more than a fresh fit}
    \STATE $Q \leftarrow \emptyset$,\quad
           $U \leftarrow \textsc{Fit}(A,E_t)$
           \algcomment{refit}
  \ENDIF
  \STATE $Q \leftarrow [\,Q,\ U^{\top}]$
         \algcomment{augment}
\ENDIF
\STATE $P \leftarrow AQ^{\top}$
\algcomment {encode $P$ with HEVC and store the new basis}
\end{algorithmic}
\end{algorithm}

\begin{figure*}[t]
\centering
\includegraphics[width=\linewidth]{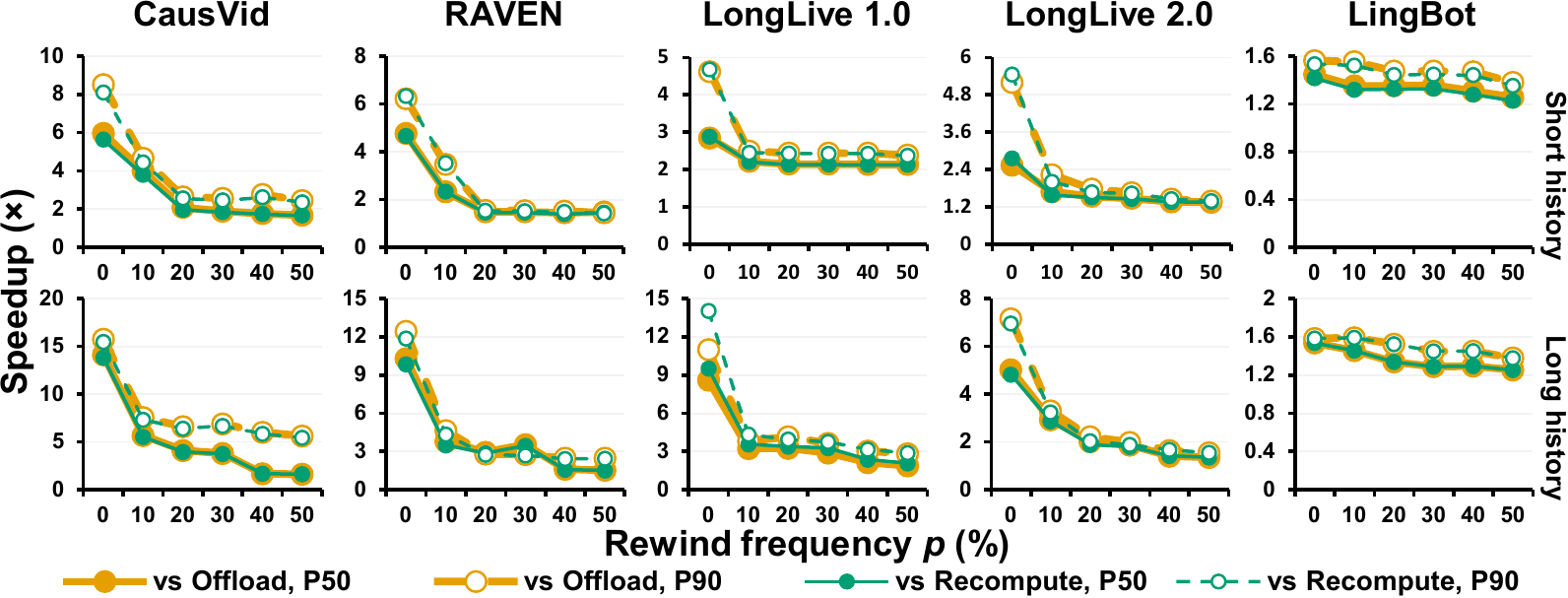}
\vspace{-6mm}
\caption{\systemname's speedup (P50 and P90) over baselines under various rewind frequency $p$.}
\label{fig:frequency_sweep}
\end{figure*}

\section{Inter-chunk Basis Reuse}
\label{s:inter_chunk_basis_reuse}

\noindent\textbf{Basis reuse and augmentation.}
For each layer and KV type, a new chunk first tests the
basis inherited from its parent. Rather than comparing two independently fitted bases, \systemname directly measures how much of the new matrix $A$ the inherited basis $\tilde Q$ captures as follows:
\begin{equation}
  P=A\tilde Q^\top,\qquad
  R=A-P\tilde Q,\qquad
  e=1-\frac{\|R\|_F^2}{\|A\|_F^2}
  \label{eq:certify}
\end{equation}
If $e\ge E_t$, the chunk reuses the shortest prefix of $\tilde Q$ that meets the target and stores its own encoded coefficients, without adding basis directions. Otherwise, the residual $R$ contains the information missing from the inherited basis. To reach $E_t$, the additional directions must capture at least $(E_t-e)\|A\|_F^2$ energy, or a fraction $(E_t-e)/(1-e)$ of the residual energy. We therefore fit the residual with this target:
\begin{equation}
  U=\textsc{Fit}\left(R,\frac{E_t-e}{1-e}\right)
  \label{eq:augment}
\end{equation}
The new directions are orthogonalized against $\tilde Q$
and appended to it. The chunk stores coefficients for the extended basis, but only the newly added basis directions.
The resulting representation is checked against the
same energy target as an independently fitted one.
Algorithm~\ref{alg:commit} describes this whole process.
On an RTX PRO 6000 GPU, the projection-energy test takes only 0.11~ms.

\noindent\textbf{When to refit.}
Repeated augmentation can increase the number of
coefficients needed by subsequent chunks. \systemname therefore fits a fresh basis when its estimated storage cost is lower than that of augmentation, accounting for both coefficients and newly stored basis directions. Specifically, it estimates the fresh rank from the layer's most recent fit, avoiding a separate decomposition just to make this decision. Large changes in content can favor refitting by requiring many additional directions, without a separate scene-change detector. A refit replaces the active basis for subsequent chunks; earlier nodes retain their original basis references. Likewise, a fork inherits the basis at its branch point and can extend or replace it independently, without modifying the other branch.

\noindent\textbf{Random access.}
Although chunks share basis directions, in \systemname each chunk's coefficients are encoded as a separate codec stream.
This allows a chunk to be restored from its coefficients
and referenced basis without decoding preceding chunks'
coefficient streams, supporting random access for
rewind and fork.

\section{Corpus Construction}
\label{app:corpus}
We construct our content corpus by extending MemFlow
interactive multi-prompt video generation benchmark~\cite{ji2025memflow},  which contains 100 narrative scripts with six successive prompts each.
We extend each script to 20 consecutive prompts, and author 400 additional 20-prompt narratives with GPT-6 Astra.
For every narrative, we also write two alternative continuations after each of five
prefixes that end at the 4th, 8th, 12th, 16th, and 20th prompt.
The resulting \emph{Synthetic-500} corpus contains 500 narratives, 10,000 main-path
prompts, and 5,000 alternative prompts.
Since users can rewind to arbitrary earlier chunks and continue sessions beyond 20 prompts, we author 1,328 additional prompts for branches they visit.
For each narrative, we author prompt-alternative branches
after each prefix ending at prompt 4, 8, 12, 16, or 20.
Each branch forms a consecutive alternative
continuation. 
For LingBot~\cite{lingbot}, it requires interactive control instead of new prompts.
We use the randomly generated camera/action sequences as the narratives.
The request arrival pattern follows a Poisson distribution.

\section{Workload Sensitivity}
\label{ss:workload_sensitivity}

We vary the rewind frequency $p$ from 0\% to 50\%  in 10-percentage-point increments, where $p$ controls the fraction of requests that resume from a historical chunk rather than extend the latest chunk. 
Figure~\ref{fig:frequency_sweep} reports P50 and
P90 next-chunk latency speedups over \emph{Offload} and
\emph{Recompute} for all five models under short-history (20 chunks) and
long-history (40 chunks) configurations. 
Within each configuration, session count, request load, and memory budget remain fixed.

\systemname outperforms both baselines throughout the sweep.
Its relative advantage generally decreases as rewinds become
more frequent. Frequent rewinds can limit forward-history growth,
reducing the baselines' historical KV reconstruction costs and
narrowing the latency gap. For CausVid under the long-history
configuration, P50 speedups over both baselines decrease from 14$\times$ at $p=0$ to
1.6$\times$ at $p=0.5$,
while P90 speedups remain above 5$\times$ at $p=0.5$. Across all
models and both configurations, P50 speedups at 50\% rewind
frequency remain 1.2--2.1$\times$.
The degree of sensitivity varies by model. For the sliding-window attention-based
LongLive~1.0, the short-history curves remain relatively
flat between 10\% and 50\%. LingBot's gains stay around
1.2--1.6$\times$ across the sweep. LingBot's expensive denoising step in its next-chunk
latency limits the improvement space for faster historical KV recovery.
 
Overall, \systemname benefits both ordinary per-session
continuation ($p=0$) and interactive rewind workloads ($p>0$).

\section{Video Frame Comparison}
\label{app:video_frame}
\begin{figure*}[t]
\centering
\includegraphics[width=\linewidth]{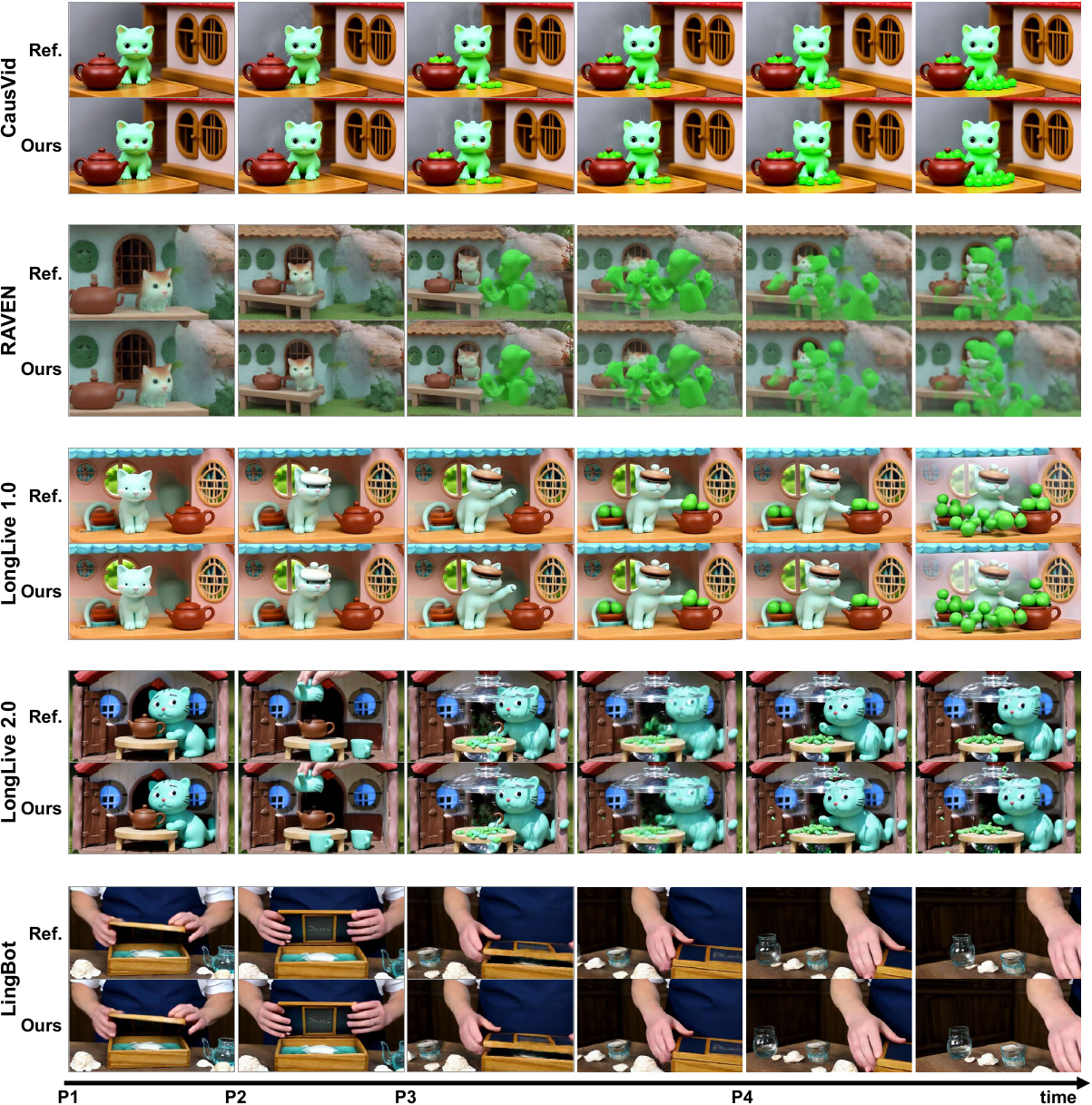}
\vspace{-7mm}
\caption{Comparison between reference videos without KV compression and matched videos generated with \systemname.}
\vspace{-6mm}
\label{fig:video_comparison}
\end{figure*}

Figure~\ref{fig:video_comparison} presents sampled frames from
reference videos generated without KV compression and matched
videos generated with \systemname.
For CausVid, RAVEN, LongLive~1.0, and LongLive~2.0, we use the
same four-prompt narrative: \textbf{P1:} A mint clay cat sets a brown teapot on a low table
inside a tiny tea house with round windows.
\textbf{P2:} The mint clay cat lifts the teapot lid, releasing
a soft steam curl above three empty cups on the table.
\textbf{P3:} The clay cat catches steam beneath a clear lid,
where tiny green shapes gather over the first empty cup.
\textbf{P4:} The mint clay cat taps the clear lid with one paw,
and the tiny green shapes scatter in a swirl before drifting
back together beneath it.
For LingBot, we use a fixed scene prompt followed by a
camera-control action sequence. The scene prompt is:
A man with a shaved head and navy apron opens a shallow wooden
shadow box on a studio table beside clean shells and sea glass.
Rather than changing the text prompt between rounds, subsequent
actions control the camera movement.

Across the displayed examples, the sampled frames generated with
\systemname closely resemble their uncompressed references,
including those from LingBot's camera-controlled sequence.
These qualitative comparisons complement the quantitative quality
evaluation and illustrate that \systemname can preserve visual
fidelity while compressing historical KV state.

\end{document}